\documentclass[acmsmall]{acmart} 

\usepackage{amsmath,amsfonts}
\usepackage{graphicx}
\usepackage{textcomp}
\usepackage{xcolor}
\usepackage{times}
\usepackage{array}
\usepackage{multirow}
\usepackage{mathrsfs}
\usepackage{soul}
\usepackage{url}
\usepackage{graphicx}
\usepackage{amsthm}
\usepackage{booktabs}
\usepackage[ruled,vlined]{algorithm2e}
\usepackage{bbm}
\usepackage{dsfont}
\newcommand{\ud}[1]{\underline{#1}}
\newcommand{\udb}[1]{\underline{\bf{#1}}}

\AtBeginDocument{%
  }

\setcopyright{acmcopyright}
\copyrightyear{2018}
\acmYear{2018}
\acmDOI{XXXXXXX.XXXXXXX}

\acmJournal{JACM}
\acmVolume{37}
\acmNumber{4}
\acmArticle{111}
\acmMonth{8}
\begin{document}
\title{From Asset Flow to Status, Action and Intention Discovery: \\Early Malice Detection in Cryptocurrency}

\author{Ling Cheng}\affiliation{%
  \institution{Singapore Management University}
  \country{Singapore}}
\author{Feida Zhu}
\affiliation{%
  \institution{Singapore Management University}
  \country{Singapore}}
\author{Yong Wang}
\affiliation{%
  \institution{Singapore Management University}
  \country{Singapore}}
\author{Ruicheng Liang}
\affiliation{%
  \institution{Hefei University of Technology}
  \city{Heifei}
  \country{China}}
\author{Huiwen Liu}
\affiliation{%
  \institution{Singapore Management University}
  \country{Singapore}}

\renewcommand{\shortauthors}{Cheng et al.}

\begin{abstract}
Cryptocurrency has been subject to illicit activities probably more often than traditional financial assets due to the pseudo-anonymous nature of its transacting entities. An ideal detection model is expected to achieve all three critical properties of (I) early detection, (II) good interpretability, and (III) versatility for various illicit activities. However, existing solutions cannot meet all these requirements, as most of them heavily rely on deep learning without interpretability and are only available for retrospective analysis of a specific illicit type.
To tackle all these challenges, we propose Intention-Monitor for early malice detection in Bitcoin (BTC), 
where the on-chain record data for a certain address are much scarcer than other cryptocurrency platforms.

We first define asset transfer paths with the Decision-Tree based feature Selection and Complement (DT-SC) to build different feature sets for different malice types. Then, the Status/Action Proposal Module (S/A-PM) and the Intention-VAE module generate the status, action, intent-snippet, and hidden intent-snippet embedding. With all these modules, our model is highly interpretable and can detect various illegal activities. Moreover, well-designed loss functions further enhance the prediction speed and model's interpretability.
Extensive experiments on three real-world datasets demonstrate that our proposed algorithm outperforms the state-of-the-art methods.
Furthermore, additional case studies justify our model can not only explain existing illicit patterns but can also find new suspicious characters.

\end{abstract}




\maketitle

\section{Introduction}
\label{sec:intro}
Cryptocurrency has emerged as a new financial asset class with ever-increasing market capital and importance. 
Together with the growing popularity comes a wide range of cybercrimes~\cite{A4_,A8_,A16_} including hacking, extortion \cite{A2_,A6_} and money laundering~\cite{A3_,A7_,A15_,A17_}. Customary in this domain, these criminal behaviors are referred to as \emph{malicious} behaviors, and the addresses committing these behaviors as \emph{malicious addresses} since each transacting entity in cryptocurrencies is represented as an anonymous address instead of an account bound with a verified identity. 
The detection and diagnosis of malicious addresses in cryptocurrency transactions present greater challenges than fraud detection in the traditional financial world for the following three distinguishing characteristics of cryptocurrency. 

\begin{figure}
	\centering
	\vspace{-0ex}
	\includegraphics[width=.9\columnwidth, angle=0]{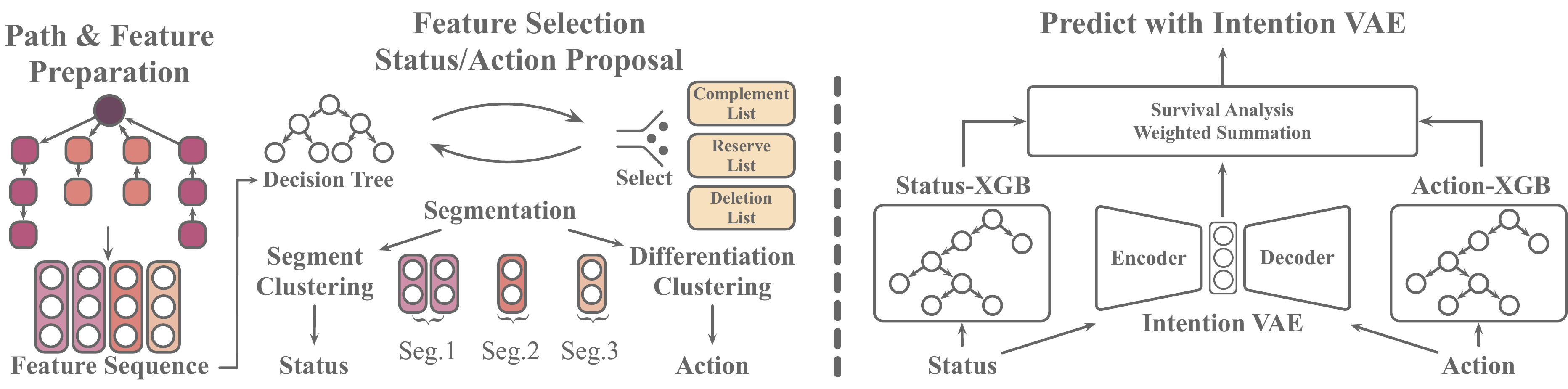}
	\vspace{-0ex}
	\caption{An overview of our Intention Monitor. 
                 After extracting address and path features, the model will select and complement the most significant features. 
                 Then, a dynamical segmentation module splits the observation period into several coherent segments and maps the segments and their differentiation to a set of global statuses and actions through clustering.
                 Finally, the Intention VAE module weights the contribution of status/action and then fine-tunes the weighted summation of two XGB models' predictions.}
	\label{fig:intro_pipeline_0}
	\vspace{-2ex}
\end{figure}

\begin{itemize}
\item
\textbf{Early detection is all that matters}. Unlike all other traditional financial assets, cryptocurrencies are traded at a 24/7, never-sleeping pace. Most malicious behaviors last for a short duration, measured only by hours, and will have already inflicted the damage before the associated malicious addresses are forever abandoned if they are not detected in the early stage. Moreover, due to the decentralized nature of cryptocurrency's peer-to-peer transactions, retrospective analysis and identification provide little value as financial losses are almost impossible to be held back and recover once the perpetration is complete. 
This challenges most existing graph-based methods as transaction graphs~\cite{A5_,A12_,A13_} needed by these methods must be sufficiently large to provide useful structural information~\cite{A1_}. In most cases, the time it takes to form such a transaction graph is much too long to respond effectively to malicious behaviors in action. Besides, these methods are usually computationally expensive and time-consuming for early-stage detection. 

\item
\textbf{Type-specific features are not versatile enough for malicious behavior detection}. The types of malicious behaviors in cryptocurrencies are increasingly diverse, complex, and constantly evolving, ranging from bitcoin-based scams to darknet markets and modus operandi hacking attacks~\cite{WY_rect_1}. The characteristics of malicious behaviors also vary a lot across different types.
Manually-engineered features from specific malicious behaviors cannot be generalized to other types and unknown ones, let alone apply to other cryptocurrencies with a complex heterogeneous relationship in general~\cite{A9_, TKA4_}. Although some studies \cite{R12_1} can detect categories of malicious activities, they are still only available for post-hoc analysis and invariably require a full-history feature observation, which is consequentially scarce at the early stage of these fraud activities. Thus, they cannot be directly deployed to detect illicit activities at the early stage. A more general class of features that capture more fundamental characteristics of malicious behaviors across different types is required to achieve the desired versatility.

\item
\textbf{Interpretability is essential}. Many malicious behaviors among the cryptocurrency platforms are packaged as commercial projects to lure victims into investing. Investors must be able to investigate and tell real creditable projects from fraudulent ones. However, most detection methods nowadays hardly offer insights into the model's predictions~\cite{A10_}. In particular, most models tend to improve recall for better safety and work appropriately for the surveillance department. However, it may increase the risk of missing investment opportunities for common investors. From the perspectives of regulators and investors, model interpretability that offers a deeper understanding of the underlying \emph{intention} behind malicious behaviors is crucial for correctly assessing and identifying malicious behaviors. 
\end{itemize}

Moreover, among all cryptocurrency platforms, Bitcoin (BTC) has the largest volume, while the on-chain record data for a certain address are much scarcer than other popular platforms (e.g., ETH, EOS with smart contracts). Thus, methods proposed on BTC are compatible with those on other cryptocurrency platforms.
To address the above-mentioned challenges in cryptocurrency platforms, we propose Intention Monitor on BTC, an early malice detection system based on the notion of \emph{asset transition paths}. The essential idea is based on the fact that, no matter which malicious behavior, the ultimate motivation and damage are reflected in the significant asset transition between innocent addresses and malicious ones. Patterns extracted from significant asset transition would therefore reveal the intention of malicious behavior across different types. Due to the generality of our asset transition paths, our Intention Monitor is potentially applicable and compatible with domain-specific techniques in other cryptocurrencies.
On a high level, our solution progresses in the following four stages:

\noindent\textbf{(I) Feature formation.} As shown in Fig.\ref{fig:intro_pipeline_0}, firstly, long-term (LT) and short-term (ST) transition paths, the features of the greatest descriptive power and versatility for early-stage malice, would be generated to capture the transaction patterns for both LT and ST transition structures.

\noindent\textbf{(II) Feature selection and complement.} Secondly, a Decision-Tree based Feature Selection and Complement (DT-SC) module would identify features of the best discriminative power for different malicious behavior types.

\noindent\textbf{(III) Temporal assembly and semantic mapping.} Status/Action Proposal Module (S/A-PM) dynamically assembles the observation period into several temporally coherent segments.  
It then maps all temporally coherent feature segments to a global status set through clustering.
Also, the differentiations between consecutive segments are mapped to global action clusters. 
These statuses and actions constitute the semantic units, and a status-action tuple (status, action) is used to denote the corresponding intent-snippet at the same time step.

\noindent\textbf{(IV) Model training with intention motif as prediction witness.} Status-based and action-based XGB models are trained to give backbone predictions. The hidden intent-snippet embedding will be proposed by the Intention-VAE module to weight the contribution of the two backbone predictions. Furthermore, these hidden snippet embeddings are used to fine-tune the predictions with a survival module and sequence into intention motifs which serve as a witness to the prediction result.

To summarize, the key contributions of this paper are as follows:
\begin{itemize}
    \item
     We propose a novel definition of asset transfer path, which is effective in capturing BTC transaction patterns for early malice detection and applicable to other cryptocurrencies potentially, making the versatility of the model across different malicious behavior types possible. 
    \item
     We provide good interpretability for our malice detection result with intention motif as prediction witness, which is unachievable by those entire deep-learning models. 
     This is achieved by a combination of
     (I) our DT-SC module to select features and S/A-PM to assemble the observation period and propose statuses, actions, and intent-snippets.
     (II) an Intention-VAE module which encodes intent-snippet into hidden embeddings to weight the contribution of different information dynamically, and (III) the survival module of Intention-VAE to fine-tune the backbone predictions and group intent-snippets into the sequence of intention motif. 
    \item
     We conduct extensive evaluation and perform substantially better on three malicious data sets than the state-of-the-art. Furthermore, we present a deep-dive case study on the 2017 Binance hack incident to illustrate the corroborating transaction patterns and unexpected hidden insights for early-sate malice detection that are otherwise unattainable. 
\end{itemize}

\section{Related Work}
\label{sec:related}
There are many crimes involving many addresses on cryptocurrency trading platforms. Therefore, detecting the identity information of the address is of great significance to the event's post-analysis and early prediction. Based on the types of features, we divide the existing malicious address detection methods into three categories: (1) case-related features; (2) general address features; (3) network-based features.

\subsection{Case-Related features}
Case-related features model the addresses and activities in a specific event. These detailed analyses are based on the IP addresses of object nodes, public data from exchanges, and labels from related forums. Concretely, some victims provided the criminals' addresses and the detail of the criminal cases they experienced. Except for those detailed data, the time difference between illegal transactions and the sub-structure in criminal cases is also helpful in malice-case analysis. 
Reid and Harrigan \cite{R1_} combined these topological structures with external IP address information to investigate an alleged theft of BTC.    
To extract information from social media, a transaction-graph-annotation system \cite{R2_} is presented. It matched users with transactions in darknet organizations' activities.
Similarly, by exploiting public social media profile information, in \cite{R3_}, they linked 125 unique users to 20 hidden services, including Pirate Bay and Silk Road.
Marie and Tyler \cite{R4_} presented an empirical analysis of BTC-based scams. By amalgamating reports in online forums, they identified 192 scams and categorized them.
Instead of direct numerical analysis, other prior studies \cite{R5_, R6_} detected three anomalous ``worm'' structures associated with spam transactions by visualizing the transaction data.
The case-related features are often helpful in specific case studies. 
However, most insights are only available in particular cases and can not be generalized to other issues. 
Thus, we put forward the asset transition paths, which are general in all event analyses.

\subsection{General address features}
The case-dependent criminal knowledge should be generalized to criminal patterns for a more general detection system.
Many works resort to machine learning for malicious activities and illegal address detection. The first step for a machine learning model is a feature proposal module \cite{A14_}. Since the transaction is the only possible action for a BTC address, commonly used address features majorly describe related transactions, revealing behavior preferences for the given address. 
Elli et al. \cite{R7_} proved transaction patterns such as transaction time, the index of senders and receivers, and the amount value of transactions can help reveal addresses' identity.
Francesco et al. \cite{R10_} proposed a method for entity classification in BTC. By performing a temporal dissection on BTC, they investigated whether some patterns are repeating in different batches of BTC transaction data. 
On ETH, Chen et al. \cite{R8_, R9_} extracted features from user accounts and operation codes of the smart contracts to detect latent Ponzi schemes implemented as smart contracts. Considering the intrinsic characteristics of a Ponzi scheme, the extracted features mainly describe the transaction amount, time, and count in a specific period.
Yin et al. \cite{TKA6_} applied supervised learning to classify entities that might be involved in cybercriminal activities.
Akcora et al. \cite{A2_} applied the topological data analysis (TDA) approach to generate the BTC address graph for ransomware payment address detection. 
Shao et al. \cite{TKA8_} embedded the transaction history into a lower-dimensional feature for entity recognition.
Nerurkar et al. \cite{R12_1} used $9$ features to train the model for segregating $28$ different licit-illicit categories of users.

The proposal of the address feature has significantly improved the model's generality. However, address features are challenging to characterize the behavioral characteristics of addresses immediately. Some particular patterns of capital inflow and outflow are difficult to be reflected in these characteristics (e.g., the pyramid structure in the Ponzi scheme and the peeling chain in money laundering). Moreover, the discriminative features are various from type to type. The redundant features are essential noise for detection.
Therefore, based on the asset transition path, we propose forward and backward concepts to describe the inflow and outflow of the asset. Furthermore, our DT-SC can propose the best feature sets for different types of activities.

\subsection{Network-based features}
Cryptocurrency inherently provides a transaction network between addresses. 
Besides focusing on address-level information, network-based features aim to characterize abnormal addresses from network interaction behaviors. By building an address or transaction network, graph metrics are proven powerful in detecting malicious activities.
By taking advantage of the power degree laws and local outlier factor methods on two BTC transaction graphs, Pham and Lee \cite{R13_} detected the most suspicious 30 users, including a justified theft.
Ranshous et al. \cite{R14_} analyzed the transaction patterns centered around exchanges. Their study introduces various motifs in directed hypergraphs, especially a 2-motif as a potential laundering pattern.
Wu et al. \cite{R15_} proposed two kinds of heterogeneous temporal motifs in the BTC transaction network and applied them to detect mixing service addresses. 
EdgeProp \cite{R16_}, a GCN-based model, was proposed to learn the representations of nodes and edges in large-scale transaction networks.
Lin et al. \cite{R17_} analyzed two kinds of random walk-based embedding methods that can encode some specific network features.
Weber et al. \cite{TKA9_} encodes address transaction graph with GCN, Skip-GCN, and Evolve-GCN.
Chen et al. \cite{TKA10_} proposed E-GCN for phishing node detection on the ETH platform.
By changing the sampling strategy in Node2Vec, Wu et al. \cite{TKA11_} proposed the Trans2Vec model, which can consider the temporal information.
Li et al. \cite{TKA12_} used TTAGN to model the temporal information of historical transactions for phishing detection.

Network-based methods perform well for retrospect analysis, as they encode the structural information of transaction graphs. However, in the early stages, to hide their identities, malicious addresses often transfer their asset with a chain-like structure. Moreover, the trading network is often too small to form a discriminative topological structure.
Also, these methods may lead to over-smoothing issues and the dilution of the minority class~\cite{E9_} under the data-unbalanced setting.

\section{Problem Formulation}
\label{sec:prob_form}
\subsection{Problem Definition}
\label{sec:prob_def}
%
 
\begin{figure}
	\centering
	\vspace{-0ex}
	\includegraphics[width=.9\columnwidth, angle=0]{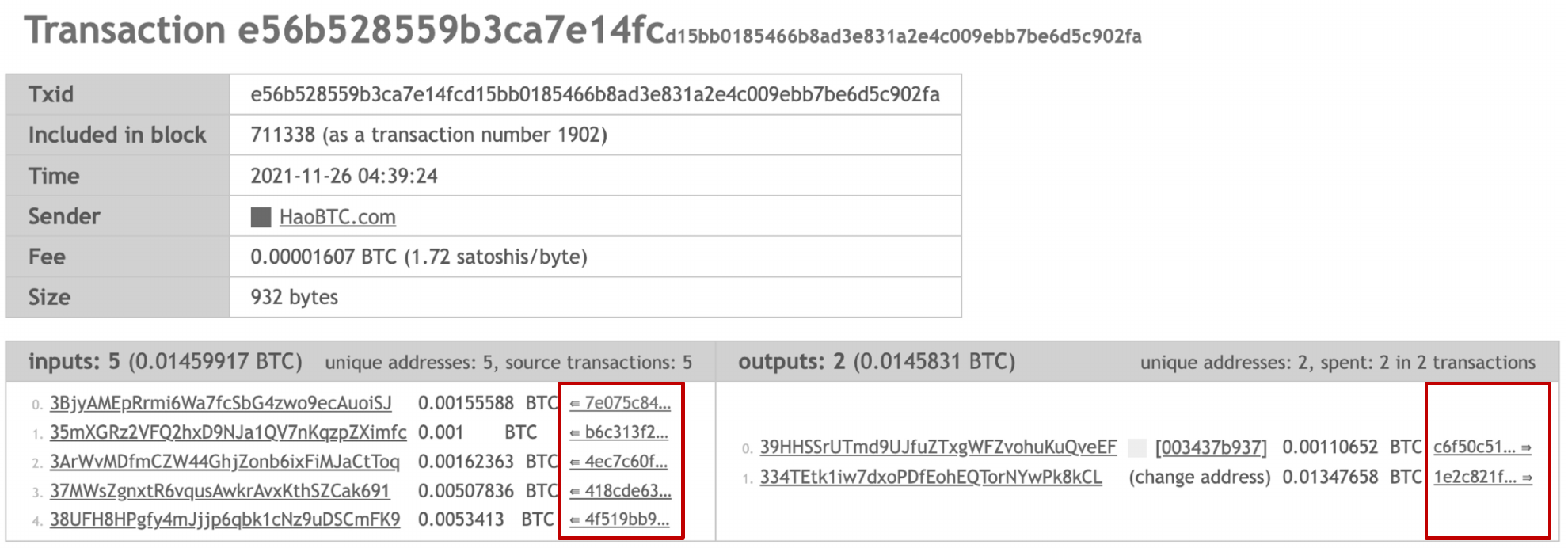}
	\vspace{-0ex}
	\caption{A BTC transaction example. The strings on the red boxes stand for $5$ input transactions and $2$ output transactions.}
	\label{fig:tx_exaple}
	\vspace{-2ex}
\end{figure}

In examining each BTC transaction, denoted as $tx$, we break down its input transaction set $I$= $\{i_1, i_2, \dots, i_{|I|}\}$ and output transaction set $J$= $\{ j_1, j_2, \dots, j_{|J|}\}$. The transaction $tx$ accounts for the redistribution of tokens between sets $I$ and $J$. To visualize it, think of incoming tokens pouring into a reservoir before being allocated to the outgoing transactions based on predetermined ratios. As shown in Fig.~\ref{fig:tx_exaple}, there are five input transactions and two output transactions in this example\footnote{https://www.walletexplorer.com/txid/e56b528559b3ca7e14fcd15bb0185466b8ad3e831a2e4c009ebb7be6d5c902fa}.
However, there is no explicit record of the number of tokens moving from an input $i$ to an output $j$. This necessitates the creation of a complete transaction bipartite graph for this $tx$, ultimately leading to the generation of $|I|\times|J|$ transaction pairs. 
Put simply, a single transaction houses $|I|\times|J|$ transaction pairs within it.
%

By the $t_m$-th time step, let $D_{t_m}$=$\{d^i_{t_m}\}_{i=1}^N$=$\{(l^i, T_{in, t_m}^{i}, T_{out, t_m}^{i})\}_{i=1}^N$, where $l^i \in \{0, 1\}$ is the label of address $i$, $0$ and $1$ stand for regular and malicious addresses respectively.
$T_{in, t_m}^{i}$=$[tx_{in, 1}^{i}, tx_{in, 2}^{i}, \dots tx_{N_{in,t_m}}^{i}]$ are transactions where address $i$ acts as the input address, and $T_{out, t_m}^{i}$ is the transaction set where address $i$ acts as the output address by the $t_m$-th time step.
For ease of understanding, we denote these two transaction sets as the $receive$ set and $spend$ set, respectively.

\vspace{0.2cm}
\noindent\textbf{Early Malicious Address Detection (EMAD).}
Given a set of addresses $A$, and 
$D_{t_m}$ at $t_m$-th time-step, 
the problem is to find a binary classifier $F$ such that 
\begin{equation}
F(d^i_{t_m})=
\begin{cases}
1& \text{if address $i$ is illicit}\\
0& \text{Otherwise}
\end{cases}.
\end{equation}
In the early detection task, we require the prediction to be consistent and predict the correct label as early as possible.
We denote the confident time as $t_{c}$, where all classifier's predictions $F$ after $t_{c}$ are consistent. 
The smallest $t_{c}$ is denoted as $t_{f.c}$. 
We aim to train a classifier to predict the correct address's label with the smallest $t_{f.c}$.

\subsection{Solution Overview}
\label{sec:solu_overview}
%
Inspired by prior research on illegal activity intention encoding \cite{M1_, M2_}, we develop a novel solution, \emph{Intention Monitor}, for the early detection of malicious addresses.

In particular, as shown in Fig.~\ref{fig:asset_trans_path}, we propose asset transfer paths to describe the transition patterns. 
These asset transfer paths can essentially capture the transaction characteristics and address intentions by tracing the source and destination of every related transaction.

Next, we put forward a Decision-Tree based Feature Selection and Complement model (DT-SC) where a decision tree model is deployed to filter and complement the most significant features for different types of malicious behaviors.
Based on the features selected by DT-SC, 
the Status/Action Proposal Module (S/A-PM) divides the observation period into several segments dynamically. 
Then, S/A-PM clusters all the addresses' segment representations and presents a set of global status representations.
The global action representations are proposed similarly based on the differentiation between two consecutive segment representations.
Each segment now has global status, action, and the corresponding intent-snippet, which can explain the behavioral intention of a given address.
Based on the status and action vectors, the status and action XGB models are trained to predict status-based and action-based
predictions.

Finally, we build \emph{Intention-VAE}, an efficient early malicious address detection framework.
The framework can 
(1) comprehensively encodes the relationship between status and action to generate the hidden intent-snippet embedding,
(2) dynamically weights the contribution between the status and action XGB models,
(3) fine-tune the weighted backbone predictions and group intent-snippets into the sequence of intention motifs.
In the subsequent sections, we introduce asset transfer path, DT-SC with S/A-PM, and Intention-VAE in Sections \ref{sec:asset_trans_path}, \ref{sec:DT_SA_with_SA_PM}, and \ref{sec:intention_vae}, respectively.

\section{Asset Transfer Path}
\label{sec:asset_trans_path}
\begin{figure}
	\centering
	\vspace{-0ex}
	\includegraphics[width=.9\columnwidth, angle=0]{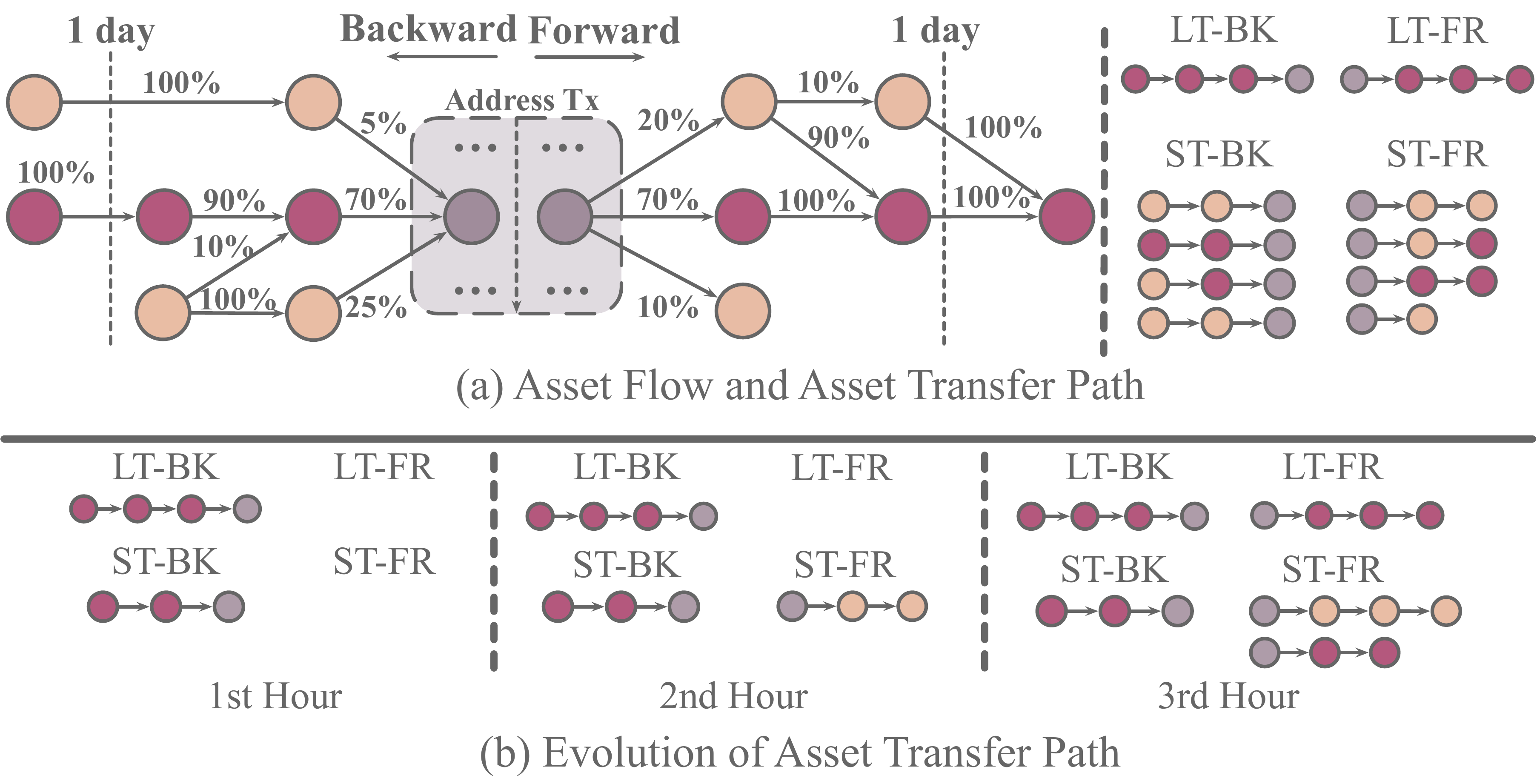}
	\vspace{-0ex}
	\caption{(a) Address transaction flow and asset transfer paths. 
                 Numbers are the amount proportions to the destination/source node. 
                 $LT$, $ST$, $FR$, and $BK$ are long-term, short-term forward, and backward, respectively. 
	          (b) Evolution of asset transfer path. 
                New asset transfer paths are generated if the address participates in new transactions. 
                Forward paths can be extended if the asset continues to flow in the next period.}
	\label{fig:asset_trans_path}
	\vspace{-2ex}
\end{figure}

At the early stage of malicious behaviors, 
the address-based network has not grown to the size for a credible prediction.
Instead, the transaction flow can provide critical information during this period.
We design asset transfer paths that consist of significant transactions for the EMAD task.

As mentioned in Sec.~\ref{sec:prob_form}, there are $|I|\times|J|$ transaction pairs in one BTC transaction.
However, not all transaction pairs are helpful for malicious address detection. 
Those important transactions typically constitute a significant portion of the entire transaction amount.
We call such transactions \emph{significant transactions}. 
%
As illustrated in Fig.~\ref{fig:asset_trans_path}, each node represents a transaction. 
The  nodes inside the dashed gray box stand for transactions that the given address participates in.
The left node is the address's \emph{receive transaction}, where the address receives tokens from other transactions.
In this single transaction example, there are three inputs, 
contributing $5\%$, $70\%$, and $25\%$ to the total transaction amount. 
Similarly, the right node is the address's \emph{spend transaction}, where the address transfers its tokens to the outputs.
This transaction involves multiple outputs (with a distribution of $20\%$, $70\%$, and $10\%$ as in this example).
We then propose influence and trust transaction pairs.
\subsection{Influence transaction pair} Given an input set $I$= $\{i_1, i_2, \dots i_{|I|}\}$ 
to an output $j$ and the transaction pair set is $\{{I\rightarrow j}\}$, i.e., $\{{I\rightarrow j}\}$=$\{(i_1,j),(i_2,j),\ldots,(i_{|I|},j)\}$, we define \emph{influence transaction pair} as follows: 
Given an influence activation threshold $\theta$,
$(i_{k},j)$ is called an \emph{influence transaction pair} for transaction $j$,
if there exists a $k$ ($1\le k\le |I|$) such that the amount of transaction pair $(i_{k},j)$ contributes to at least a certain proportion of the input amount of transaction $j$, i.e, $\hat{A}(i_{k},j) \ge \theta \times \hat{A}(\{{I\rightarrow j}\})$, where $\hat{A}(\cdot)$ denotes the amount of a transaction pair or the sum of all transaction pairs.

Given an influence transaction pair $(i_{k},j)$, we can conclude that output $j$ obtains at least a significant amount (based on the threshold) of the asset in this transaction from input $i_{k}$. Accordingly, given an \emph{receive transaction} $j$ for the given address, to trace back to the source of the asset, we proposed \emph{backward path} based on \emph{influence transaction pair}. 
Algo. \ref{alg:bk_path} gives the detail to prepare \emph{backward paths} that reveal where $j$ obtains the asset.

\subsection{Trust Transaction Pair}
In addition to tracing back to the asset source, we also need to investigate where the asset flows to, i.e., the destination of the asset transfer. To that end, we define \emph{trust transaction pair} as follows:
Given a set of outputs $J$= $\{j_1, j_2, \dots j_{|J|}\}$,
an Input $i$, and the set of all transaction pairs $\{{i\rightarrow J}\}$, 
%
for an output $j_k$ ($1\le k\le |J|$),
if the input $i$ transfers at least a certain proportion of its output amount to it, this transaction pair is called a \emph{trust transaction pair} for Input $i$.
It indicates a specific form of trust from $i$ to $j_k$ in an asset transfer. 

Given a trust transaction pair $(i,j_k)$, we can conclude that input $i$ sends a certain degree of the asset to output $j_k$. 
To trace the destinations of Input $i$, we also define \emph{forward paths} based on \emph{trust transaction pair}. 
The pipeline to construct \emph{forward path} is similar to \emph{backward path}. The only difference is the tracing direction.

 \begin{algorithm}
 \caption{Backward path preparation}
 \label{alg:bk_path}
  \DontPrintSemicolon
  \SetKwInOut{Input}{input}
  \SetKwInOut{Output}{output}
  \Input{Initial Output Tx $j^o$, Threshold $\theta$, Time Span $T_{Span}$.}
  \Output{Backward Path Set $P$.}
 
 \nl Initialize Backward Path Set: $P \gets \{[-,1,j^o]\}$;\;
 \nl Initialize Previous hop recorder: $P_{pre} \gets \{[-,1,j^o]\}$;\;
 \nl Initialize Ending Flag: $F_{end} \gets False$;\;
 \nl $j^o$'s Time: $T_{j^o} \gets $ Time of $j^o$;\;

  \nl \While{$F_{end} \ne True$}{
  \nl Current hop recorder $P_{now} \gets \{\}$;\;
  \nl $F_{end} \gets True$;\;
  \nl  \For {$p$ in $P_{pre}$}{
  \nl        $j \gets $ Output Tx $p[2]$;\;
  \nl        $I \gets $ Input Tx Set of $j$;\;
  \nl        \For {$i$ in $I$}{
  \nl                  $Prop_{i} \gets Amt_{i}/Amt_{I}$;\;
  \nl              $Score_{i} \gets Prop_{i}*p[1]$;\;
  \nl              $T_{i} \gets $ time of $i$;\;
  \nl              \If {($Score_{i} \ge \theta$ and $T_{j^o} - T_{i}\le T_{Span}$)}{
  \nl                     Append $[j, Score_{i}, i]$ to $P_{now}$;\;
  \nl                     $F_{end} \gets F_{end}$ $\&\&$ $False$;\;}}}
\nl $P_{pre} \gets P_{now}$;\;
\nl $P \gets P \cup P_{pre}$;\;
}
\nl \Return{$P$} 
\end{algorithm}

\subsection{Long-term and Short-term Path} 
For brevity, we would refer to both the \emph{backward path} (BK) and \emph{forward path} (FR) as \emph{asset transfer paths} and the activation threshold in both directions as \emph{activation threshold}.
To delineate the transaction patterns of an address at both macro and micro levels for both \emph{backward path} and \emph{forward path}, 
we define two kinds of time spans, i.e., long-term and short-term.

Long-term (LT) asset transfer paths have a larger maximum observation period and higher activation threshold, as they are designed to find the transaction's major asset source. 
Short-term (ST) asset transfer paths have a shorter observation period and lower activation threshold. 
They describe the transition pattern and structure (e.g., pyramid-shaped, pulse-shaped, and spindle-shaped) within a short period. 


%

\section{Feature Selection and Complement \& Status Proposal Module}
\label{sec:DT_SA_with_SA_PM}
\begin{algorithm}
\caption{DT-Based Feature Selection and Complement}
\label{alg:DT_SC}
  \DontPrintSemicolon
  \SetKwInOut{Input}{input}
  \SetKwInOut{Output}{output}
  \Input{Initial feature list $F^i$, Threshold $\theta$.}
  \Output{Complement, Reserve, and Delete lists $F_C, F_R, F_D$.}
  \nl Complement and Delete feature list: $F_C, F_D \gets \{\}$;\; 
  \nl Reserve feature list: $F_R \gets F^i$;\;
  \nl Average performance score: $s_{p}^{A} \gets 0$;\; 
  \nl Best average performance score: $s_{p}^{B,A} \gets 0$;\; 

  \nl \While{$s_{p}^{A} \geq s_{p}^{B,A}$}{
   \nl $s_{p}^{B,A} \gets s_{p}^{A}$;\; 
   \nl $F^{tmp}_{C}, F^{tmp}_{R}, F^{tmp}_{D} \gets F_{C}, F_{R}, F_{D}$;\; 
   \nl Average performance score $s_{p}^{A} \gets 0$;\; 
   \nl Best performance score $s_{p}^B \gets 0$;\; 
   \nl \For{$idx \gets 1$ to $10$}{
       \nl $DT_{idx}, s_{p, idx} \gets$ Train\&Test DT($F_C, F_R, F_D$);\; 
       \nl $s_{p}^{A} \mathrel{+}= s_{p, idx}/10$;\; 
       \nl \If{$s_{p, idx}>S^B_p$}{
            \nl Update($F^{tmp}_{C}, F^{tmp}_{R}, F^{tmp}_{D}$);\;}}

\nl \If {($s_{p}^{A} \geq s_{p}^{B,A}$)}{
\nl $F_C, F_R, F_D \gets F^{tmp}_{C}, F^{tmp}_{R}, F^{tmp}_{D}$;\; 
}}
\nl \Return{$F_C, F_R, F_D$}
\end{algorithm}

\begin{figure}
	\centering
	\vspace{-0ex}
	\includegraphics[width=.9\columnwidth, angle=0]{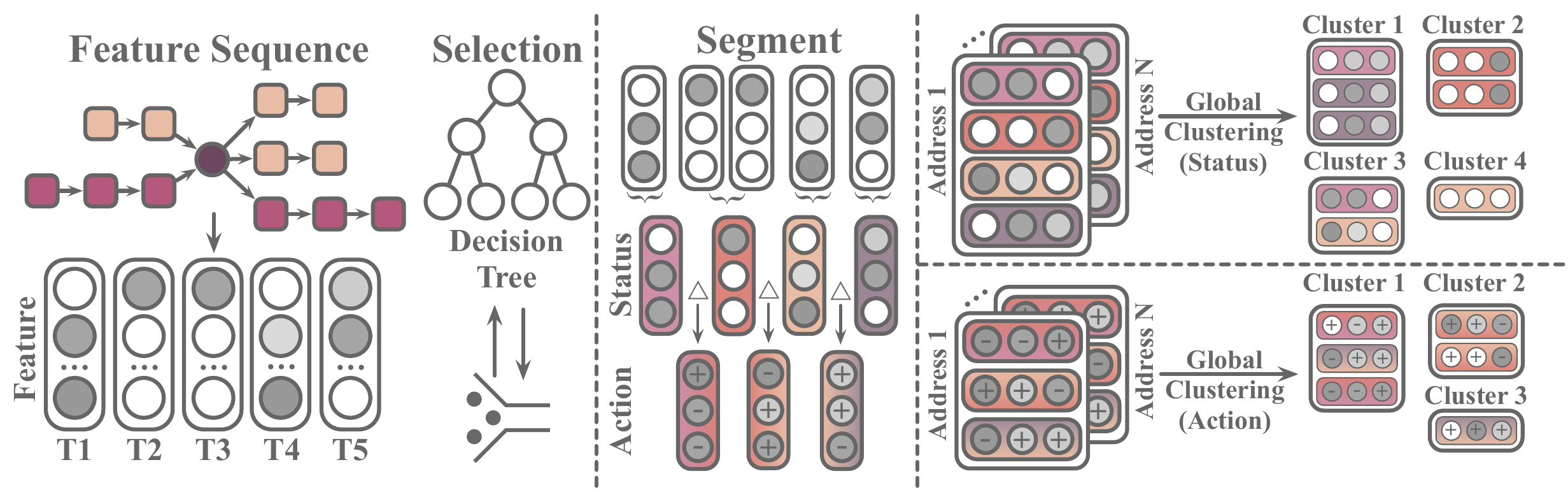}
	\vspace{-0ex}
	\caption{An overview of Decision-Tree based feature Selection and Complement (DT-SC) and Status/Action Proposal Module (S/A-PM). 
                 After extracting address and path features, The DT-SC will filter and complement the most significant features. 
                 Then, the S/A-PM splits the observation period into several coherent segments and maps the segments and their differentiation to a set of global statuses and actions through clustering.}
	\label{fig:intro_pipeline_1}
	\vspace{-2ex}
\end{figure}

In this section, we describe the process of Decision-Tree based Feature Selection and Complement (DT-SC) and Status/Action Proposal Module (S/A-PM). 
We first introduce the address and asset transfer path features. 
Then, we elaborate on DT-SC that filters, and complements the features for different malicious activities.
The status and action sequences are necessary to understand the address's intention. 
To fetch these sequences, we deploy S/A-PM to split the observation period into segments and cluster them to generate global status and actions.
The overview of DT-SC and S/A-PM is shown in Fig.~\ref{fig:intro_pipeline_1}.

\subsection{Address \& Transaction Features}
\begin{table}[]
\centering
\fontsize{8}{9}\selectfont 
\caption{Feature counting and explanation.}
\begin{tabular}{ccccc}
\toprule
Feature Type & Aspect     & Feature      &Num. &Complement  \cr
\midrule 
\multirow{11}{*}{{Address}}        & Balance            & Balance   & 1        & None   \cr
            \cmidrule(l{0em}r{0em}){2-5} 
            & \multirow{2}{*}{Tx Count}      & Number of spend (receive) tx (by now/recent one hour) & 4 & None   \cr
            &        & Ratio of spend tx to recieve tx (by now/recent one hour)   & 2        & None   \cr
            \cmidrule(l{0em}r{0em}){2-5} 
            & Tx Frequency        & Max spend (receive) tx number per hour      & 2        & None    \cr
            \cmidrule(l{0em}r{0em}){2-5} 
            & Abnormal Tx        & Number of spend (receive) tx   with 0 amount  & 2        & None    \cr
            \cmidrule(l{0em}r{0em}){2-5} 
            & \multirow{2}{*}{Temporal Info} & Time of max hourly spend (receive) tx number            & 2        & None    \cr
            &        & Time difference between max hourly spend and receive    & 1        & None    \cr
            \cmidrule(l{0em}r{0em}){2-5} 
            & Activity           & Active hour number and active rate          & 2        & None    \cr
\midrule 
\multirow{6}{*}{{Path}}   & Path-Count         & Path number           & 1        & None    \cr
            \cmidrule(l{0em}r{0em}){2-5} 
            & Path-Length        & Hop (height)-length        & 2        & Min,Max,Std         \cr
            \cmidrule(l{0em}r{0em}){2-5} 
\multirow{2}{*}{{(LT/ST)-(BK/FR)}}   & Tx Amount          & Max (min) input (output) amount   & 4        & Min,Max,Std         \cr
            \cmidrule(l{0em}r{0em}){2-5} 
            & Tx Structure       & Max (min) input (output) tx number            & 4        & Min,Max,Std         \cr
            \cmidrule(l{0em}r{0em}){2-5} 
            & Connectivity       & Path's max (min) activation score & 2        & Min,Max,Std       \cr 
\bottomrule
\label{tab:feat_explain}
\vspace{-2ex}
\end{tabular}
\end{table}

Following \cite{R11_, R12_1, R12_2}, we also use the address features to characterize the address's behaviors.
As shown in Table~\ref{tab:feat_explain},
we extracted 16 address features that characterize an address from six perspectives.
Moreover, the asset path can provide critical information. For a specific path set, we selected 13 path features from five perspectives.
Also, an address has four path sets (LT-BK, ST-BK, LT-FR, ST-FR), a path set has multiple paths,
and every path has these 13 path features. 
To characterize the overall properties of each path set,  
we calculated the maximum (max), minimum (min), average (avg), and standard deviation (std) values of every feature except the path number. 
Thus, there are 12*4+1=49 path features for a single path set. 
Since we have four path sets (LT-F, LT-B, ST-F, ST-B) (Fig.~\ref{fig:asset_trans_path}), there are 49*4=196 path features in total. 
We can characterize the early behavior of different types of addresses through these address and path features.

We believe that hundreds of extracted address and path features can summarize addresses' early behaviors from several perspectives.
But for a specific type of activity, not all features are equally helpful, and the introduction of irrelevant features can affect the model's performance (will be justified in Section~\ref{sec:feat_compare}).
Therefore, we need to select the most discriminative features from all these features.

\subsection{DT-based feature Selection and Complement}
Decision trees can partition the data based on the features that best separate the classes or target variables. During the tree-building process, features that are more informative or discriminatory tend to be selected earlier as splitting criteria. This means that important features are prioritized in the decision tree construction and thus have higher importance scores.

We develop a decision tree-based feature selection and complement module.
In this module, we have three sets: complement list, reserve list, and deletion list.
This module complements features in the complement list,
retains features in the reserve list, and deletes features in the deletion list.

In the initial round, address features and all path features' mean values are set as seed features. 
So the seed feature number is 13*4+16=68.
We feed these 68 features into the decision tree model and select the best-performing (the performing score will be elaborated in Section~\ref{sec:metrics}) model from $10$ independent training models.
We sort the model's feature importance scores and denote the maximum importance score as $s_{imp}^M$. 
Given a feature j with an importance score $s_{imp, j}$, 
if $s_{imp, j}$$\geq$$\theta$*$s_{imp}^M$, we append it into the complement list and complement it in the next round of training. $\theta$ is the complement threshold. If $0<s_{imp, j}<\theta$*$s_{imp}^M$, we append it into the reserve list, and we will retain it without complement in the following round. If $s_{imp, j}$=$0$, we append it into the deletion list, and we will delete this feature in the subsequent training process.

In the second round of training, we first complement the features in the complement list.
Here, by complement, we mean not only using the feature's mean value but also including its maximum, minimum, and standard deviation (only path features are available for complement).
Through complement, we provide the model with more details about the complement feature.
Then, we append reserve features to the input feature list without changing them.
Finally, we delete features in the delete list. 
Algo.~\ref{alg:DT_SC} shows the details of DT-SC.

\subsection{Status/Action Proposal Module}
The status and action sequences can depict the intention of the given address~\cite{M1_}, which is of great importance to the model's interpretability.  
However, if we analyze each address independently, it is difficult for the model to obtain a generalizable intention module,
the model's ability and interpretability will be reduced when predicting newly emerging malicious behaviors.
To solve these problems, we propose the Status/Action Proposal Module (S/A-PM).

\noindent\textbf{Dynamical segmentation.}
By definition, status is to describe a certain stable state.
To characterize the statuses and their evolution,
we need to split the entire observation time window into several ``state'' segments dynamically.
In every segment, we require all addresses' features to be stable enough.
Therefore, at the $J$-th time step, 
we first normalize all addresses' feature sequences along the timeline.
Take address $i$ as example,
the feature list is $[f_{1, i}, ..., f_{j, i}, ... f_{J, i}]$
where $f_{j, i}$ is the feature vector at $j$-th time step of address $i$.
And for each time step $j$, we then calculate the change ratio $C_j$.
\begin{equation}
C_j = \frac{\sum_{i=1}^{N}{\sum_{m=1}^{M}({f_{j,i}^{m}-f_{j-1,i}^{m}})/({f_{j-1,i}^{m}+{\delta}})}}{M*N},
\end{equation}
where $N$ is the data size, $M$ is the feature dimension, $delta$ is a small number in case of the divisor equals $0$.
By the $J$-th time step, the highest and current change ratio is denoted as $C_H$ and $C_J$.
If $C_J > \theta * C_H$, we think the addresses' statuses change and add $J$ to the segmentation point list.
In this manner, we can guarantee the stability of each segment.
Notice that, to build a more general segmentation strategy, the segmentation point list is shared by all addresses.

\noindent\textbf{Segment Representation.}
Then, to build a generalizable intention module, 
we need to find and represent the common ground of these segments across all addresses.
Therefore, for the $j$-th segment of address $i$, we first define its \emph{segment representation}.
The segment's beginning and end time points are denoted as $b_j$ and $e_j$ respectively,
and the corresponding feature sequence is $[f_{b_j, i}, f_{b_j+1, i}, ..., f_{e_j-1, i}, f_{e_j, i}]$.
Then the segment vector $g_{j, i}$ is calculated as the average of the feature sequence over the timeline.
Naturally, the change between $g_{j-1, i}$ and $g_{j, i}$ is redeemed as segment differentiation $d_{j, i}$.
Specially, we define $g_{0, i}$ as a full-zero vector.
Thus, for address $i$, we get a sequence of segment representations $[g_{1, i}, ..., g_{k, i}, ..., g_{K, i}]$, 
and a sequence of differentiation representations $[d_{1, i}, ..., d_{k, i}, ..., d_{K, i}]$, 
where $K$ is the segmentation number.

\noindent\textbf{Status and action clustering.}
Addresses with the same type may have similar purposes.
For example, after the Ransomware addresses are activated, 
most will wait for the victims to pay the ransom in a segment and transfer out quickly in another segment.
We can call these two segments the ``waiting segment'' and ``transfer segment'' respectively.
These semantic meanings are beneficial for understanding and interpreting the address's intention, 
and we will justify this in Section~\ref{sec:stat_and_inten}.
 
To obtain global semantic representations such as ``waiting'' or ``transfer'', 
we cluster these $N*K$ addresses' segment representations through the agglomerative clustering algorithm, a hierarchical clustering method.
The hierarchical structure provides better interpretability in further analysis.
Similarly, we cluster these $N*K$ addresses' segment differentiations to describe the general actions to obtain action clusters.
We denote each status and action cluster's centers as the status and action vectors.
Finally, for each address, we have five sequences: feature, status(vector and cluster index), and action(vector and cluster index). 
Notice that the action is calculated from the current status and previous status.

\section{Intention VAE}
\label{sec:intention_vae}
As mentioned in Section~\ref{sec:related}, tree-based machine learning algorithms have been proven powerful in related tasks about malice detection. However, they are difficult to utilize temporal patterns, which is extremely useful for depicting the address's intention, especially during the early stage with severe data scarcity. Besides, as there is no early stopping mechanism in the decision tree group, subsequent redundant noise will trigger the issue of inconsistent prediction. 
Thus, on top of tree-based algorithms (XGB as the backbone), we introduce the Intention VAE module, which can encode such temporal patterns and prevent noise with survival analysis.

\begin{figure}
	\centering
	\vspace{-0ex}
	\includegraphics[width=.9\columnwidth, angle=0]{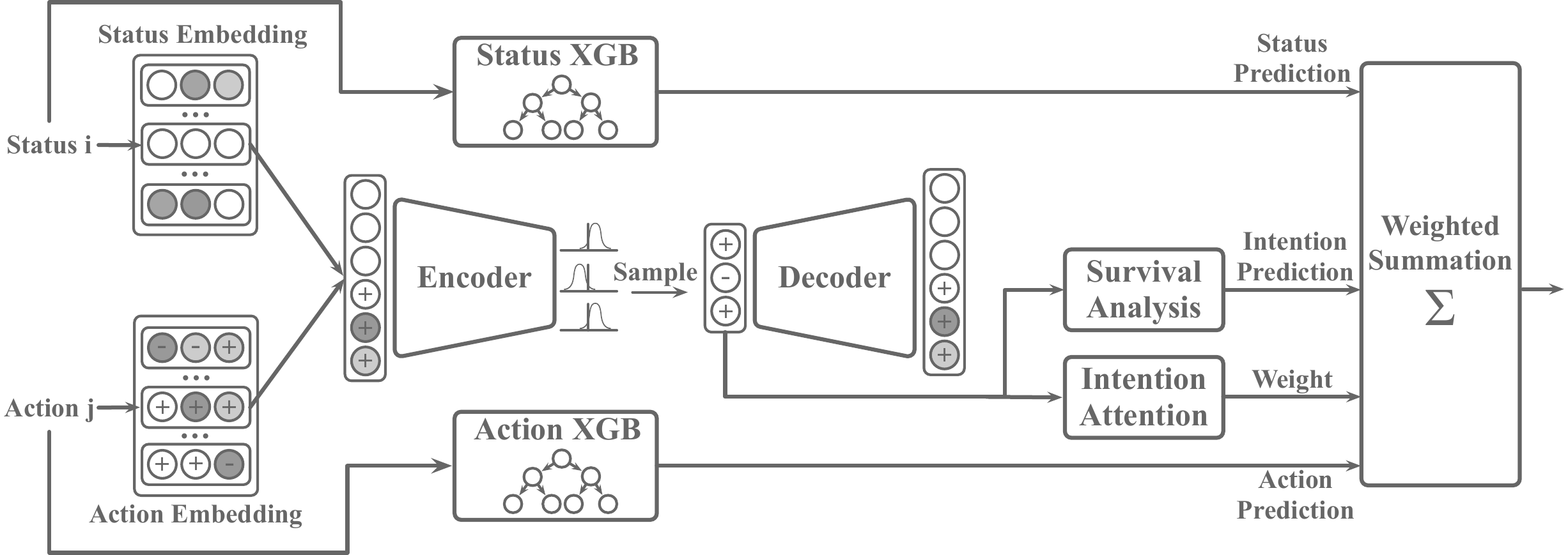}
	\vspace{-0ex}
	\caption{An overview of Intention-Variational-Autoencoder (Intention-VAE). 
                 Two XGB models are trained with status and action, respectively. 
                 At each time step, the Intention-VAE module generates hidden intent-snippet embeddings to weight the prediction of two XGB models and fine-tunes the weighted prediction.}
	\label{fig:intro_pipeline_2}
	\vspace{-2ex}
\end{figure}

\subsection{Intention Proposal}
By understanding an address's status and action, we can discern its intention. Furthermore, when an address is under a certain status, and its intention is known, we can forecast its subsequent action.
Thus, to depict this relationship, we deploy a VAE module to generate the hidden intent-snippets embeddings with the corresponding statuses and actions.
As shown in Fig~\ref{fig:intro_pipeline_2},
for address $i$ at time step $j$, we have the corresponding status index $S^{idx}_{i, j}$ and action index $A^{idx}_{i, j}$.
We transform them into learnable embedding vectors as follows:
\begin{equation}
\begin{aligned}
S^{E}_{i, j} &= \mbox{Emb}^S(S^{idx}_{i, j}),\\
A^{E}_{i, j} &= \mbox{Emb}^A(A^{idx}_{i, j}).
\end{aligned}
\end{equation}

After getting the embedding of status and action, the sampled bottleneck vector of our Intention-VAE module is regarded as the hidden intent-snippet. Through Intention-VAE, we can preserve the most critical information of status and action. Moreover, the non-linear transformation in the module can generate a more expressive representation. 
Another crucial point is that Intention-VAE can guarantee that the intention space has the properties of continuity and completeness~\cite{VAE_1}. Specifically, continuity requires two close intention points in the latent space not to give two completely different contents. As for completeness, it requires an intention point to give meaningful content once decoded.
The intention representation is calculated as follows:
\begin{equation}
\begin{aligned}
x_{i, j} &= \mbox{VAE-Encoder}([S^{E}_{i, j}||A^{E}_{i, j}]),\\
\mu_{i, j} &= W_{x{\mu}}x_{i,j}+b_{\mu},\\
\sigma_{i, j} &= W_{x{\sigma}}x_{i,j}+b_{\sigma},\\
z_{i, j} &= \mu_{i, j}+\mbox{exp}(\sigma_{i, j})\odot e, \ e\sim \mathcal{N}(0, I),\\
\hat{x}_{i, j} &= \mbox{VAE-Decoder}(z_{i, j}),
\end{aligned}
\end{equation}
where $[.||.]$ stands for concatenation, 
$\hat{x}_{i, j} \in \mathbb{R}^{d_z}$ is the output of the VAE encoder, $d_z$ is the dimension of the intent-snippet.
$z_{i, j}$ is the bottleneck vector of our Intention-VAE, also denoted as the hidden intent-snippet embedding.
Instead of decoding the interval vector directly, VAE encodes inputs as distributions.
Thus, the model proposes the mean value $\mu_{i, j}$ and stand deviation $\sigma_{i, j}$.
Our hidden intent-snippet embedding is sampled from the distribution of $\mathcal{N}(\mu_{i, j}, \sigma_{i, j})$.
However, due to the gradient descent issue of the sampling process,
the module uses a reparameterization trick to sample $e$ form $\mathcal{N}(0, \mathcal{I})$ instead of sampling $z$ directly.
$\hat{x}_{i, j}$ is the reconstruction of input data.

\subsection{Intention-Based Survival Analysis}
To enable the model to encode temporal patterns, we introduced Intention-Based Survival Analysis to Intention-VAE.
This module can accelerate the prediction speed and improve the consistency of the prediction by cutting off continuous noise.

The survival function $S(j)$ represents the probability of an event that has not occurred time step $j$, 
where an event here represents ``the given address labeled as malicious''.
Every time step, the model should give lower $S(j)$ to malicious addresses and higher $S(j)$ to legal addresses. 
The hazard function $\lambda_{j}$ is the instantaneous event occurrence rate at time $j$ given that the event does not occur before time $j$. Note that the observation time is discrete in our case. 
We denote a discrete  timestamp index as $j$, the association between $S(j)$, $\lambda_{j}$ and probability density function $f_j$ can be described as:
\begin{equation}
\begin{aligned}
\label{eq:dsicrete_survival_function}
S(j) &= P(T\geq{j}) = \sum_{k=j}^{\infty}f_k, \\
\lambda_{j} &= P(T=j|T\geq{j}) = {f_j}/S(j),\\
S(j) &= exp({-\sum_{k=1}^{j}\lambda_{k}}).
\end{aligned}
\end{equation}

To enable the model to encode temporal patterns, we first update each input with corresponding LSTM modules as follows:
\begin{equation}
\begin{aligned}
h_{i,j}^{F}, c_{i,j}^{F} &= \mbox{LSTM}^{F}([z_{i, j} || f_{i,j}], h_{i,j-1}^{F}, c_{i,j-1}^{F}),\\
h_{i,j}^{S}, c_{i,j}^{S} &= \mbox{LSTM}^{S}([z_{i, j} || S^{vec}_{i, j}], h_{i,j-1}^{S}, c_{i,j-1}^{S}),\\
h_{i,j}^{A}, c_{i,j}^{S} &= \mbox{LSTM}^{A}([z_{i, j} || A^{vec}_{i, j}], h_{i,j-1}^{A}, c_{i,j-1}^{A}).\\
\label{algo:lstm}
\end{aligned}
\end{equation}

Then, the hazard rate ${\lambda}_{j}$ and intention-based prediction in our Intention-AVE are given as follows:
\begin{equation}
\begin{aligned}
\lambda_{j} &= \sum_{{t}\in\{F,S,A\}}ln(1+exp({W_{i}h_{i,j}^{t}}))),\\
P_{i,j}^I &= [exp({-\sum_{k=1}^{j}\lambda_{k}}),\ 1-exp({-\sum_{k=1}^{j}\lambda_{k}})].
\end{aligned}
\end{equation}

\subsection{Intention Augmented Prediction Fusion}
\label{sec:Intention_fusion}
Despite the difficulty in utilizing temporal patterns, 
tree-based models can provide better stability under label unbalance settings. 
To incorporate these benefits, we build an Intention Augmented module upon the tree-based models.

First, we train two XGB models with status and action representations, namely \emph{Status-XGB} and \emph{Action-XGB}.
For address $i$ at time step $j$, based on the status vector $S^{vec}_{i, j}$ and action vector $A^{vec}_{i, j}$, we first get predictions $P_{i,j}^S, P_{i,j}^A \in \mathbb{R}^{2}$ from these two models as the backbone predictions.
Each dimension represents the probability of the corresponding class (0-normal, 1-malicious).

Both status and action can provide critical information for prediction, but their contributions may change dynamically. 
Therefore, the weights assigned in the two decision trees should also change dynamically. 
To this end, we use the Intention-Attention module to assign the prediction weights of the two decision trees. 
\begin{equation}
\begin{aligned}
a_{i, j}^{S} &= W^{a}\mbox{tanh}(W^{f, S}[f_{i, j} || S^{vec}_{i, j}]),\\
a_{i, j}^{A} &= W^{a}\mbox{tanh}(W^{f, A}[f_{i, j} || A^{vec}_{i, j}]),\\
a_{i, j}^{I} &= W^{a}\mbox{tanh}(W^{f, I}[f_{i, j} || z_{i, j}]),\\
{\alpha}_{i, j}^{(S,A,I)} &= \mbox{exp}(a_{i, j}^{t})/{\sum_{t}^{{\{S,A,I\}}}\mbox{exp}(a_{i,j}^{t})}.
\label{algo:attention}
\end{aligned}
\end{equation}

With the acceleration of the Intention-Based Survival Analysis,
the final prediction $\hat{P}_{i, j}$ for address $i$ at time step $j$ is given by:
\begin{equation}
\begin{aligned}
P_{i, j} &= {\sum_{t}^{\{S,A,I\}} P_{i, j}^{t}*a_{i, j}^{t}}, \\
\hat{P}_{i, j} &= S(j)*P_{i, j} + (1-S(j))*\hat{P}_{i, j-1}.
\end{aligned}
\end{equation}
The introduction of survival prediction analysis can correct the prediction errors caused by the lack of temporal information and group statuses into an intention sequence. 
We denote the time step when the survival probability equals $0$ as $t_{die}$, 
where the model has collected enough information to predict the malicious address's label. 
In other words, the model has figured out the address is malicious. 
Since VAE makes the distribution of each $z$'s dimension close to $\mathcal{N}(0, I)$, 
and the intention latent space is continuous and complete, 
we can binarize each dimension to index each address's intent-snippet as $I^{idx}_{i, j}$
(e.g., As shown in Fig~\ref{fig:intro_pipeline_2}, $z$'s dimension $d_z$ is $3$ , we can get $8$ intention indices, 
i.e., $1$(+,+,+), $2$(-,+,+), ..., $7$(+,-,-), $8$(-,-,-)).
Thus, we call the intent-snippet index sequence $\{I^{idx}_{i, j}\}^{t_{die}}_{j=1}$ as the intention motif of address $i$.
And this intent-snippet motif can be interpreted by the corresponding status and action sequences.

\subsection{Loss Function}
\noindent\textbf{Classification Loss}. 
For address $i$ at $j$-th time step, the early detection likelihood that this address is malicious and the negative logarithm prediction $loss^P$ are defined below:
\begin{equation}
\begin{aligned}
\hat{y}_{i,j} &= P_{i, j}[1],\\
likelihood & = (1-\hat{y}_{i,j})^{l_i}({\hat{y}_{i,j}})^{1-l_i},\\
loss^P_{i,j} & = ({l_i}-1)*\mbox{log}({\hat{y}_{i,j}}) - {l_i}*\mbox{log}(1-{\hat{y}_{i,j}}).
\end{aligned}
\end{equation}

\noindent\textbf{Intention-VAE Loss}. 
VAE is trained by maximizing the log-likelihood as follows:
\begin{equation}
\begin{aligned}
\mbox{log}P([S^{E}_{i, j}||A^{E}_{i, j}]) &= \int_{z}(q(z|[S^{E}_{i, j}||A^{E}_{i, j}])\mbox{log}P([S^{E}_{i, j}||A^{E}_{i, j}]))dz, \\
&= \int_{z}q(z|[S^{E}_{i, j}||A^{E}_{i, j}])\mbox{log}(\frac{P(z, [S^{E}_{i, j}||A^{E}_{i, j}])}{q(z|[S^{E}_{i, j}||A^{E}_{i, j}])})dz \\
&+ \int_{z}q(z|[S^{E}_{i, j}||A^{E}_{i, j}])\mbox{log}(\frac{q(z|[S^{E}_{i, j}||A^{E}_{i, j}])}{P(z|[S^{E}_{i, j}||A^{E}_{i, j}])})dz, \\
&\geq \int_{z}q(z|[S^{E}_{i, j}||A^{E}_{i, j}])\mbox{log}(\frac{P([S^{E}_{i, j}||A^{E}_{i, j}]|z)P(z)}{q(z|[S^{E}_{i, j}||A^{E}_{i, j}])})dz,
\end{aligned}
\end{equation}
where $z\sim \mathcal{N}(0, I)$ is an intermediate random variable, $\int_{z}q(z|[S^{E}_{i, j}||A^{E}_{i, j}])\mbox{log}(\frac{q(z|[S^{E}_{i, j}||A^{E}_{i, j}])}{P(z|[S^{E}_{i, j}||A^{E}_{i, j}])})dz$ equals $KL(q(z|[S^{E}_{i, j}||A^{E}_{i, j}])||P(z|[S^{E}_{i, j}||A^{E}_{i, j}]))$ which is guarantee to be positive. Thus we only need to maximize the lower bound $lb_{i, j}$ i.e., the last term in the above equation.
$lb_{i, j}$ can be reformalized as follows:
\begin{equation}
\begin{aligned}
lb_{i, j} = \int_{z}(q(z|[S^{E}_{i, j}||A^{E}_{i, j}])\mbox{log}(\frac{P(z)}{{q(z|[S^{E}_{i, j}||A^{E}_{i, j}])}})dz + 
      \int_{z}(q(z|[S^{E}_{i, j}||A^{E}_{i, j}])\mbox{log}(P([S^{E}_{i, j}||A^{E}_{i, j}]|z))dz,
\end{aligned}
\end{equation}
where the first term equals to $-KL(q(z|[S^{E}_{i, j}||A^{E}_{i, j}])||P(z))$. 
To minimize this KL divergence, we need $q(z|[S^{E}_{i, j}||A^{E}_{i, j}]))$ close to $\mathcal{N}(0, I)$.
The second term equals the negative reconstruction error in Auto Encoder.
Finally, the VAE loss $loss^V_{i,j}$ can be represented as $\sum_{d=1}^{d_z} (exp(\sigma_{i, j}^{d})-(1+\sigma_{i, j}^{d}) + (\mu_{i, j}^{d})^2)$ as indicated by~\cite{VAE_1}.

\noindent\textbf{Consistent and Early Boost Loss}.
An accurate and reliable model should provide a consistent prediction.
For an ideal model, the current prediction should be consistent with the previous prediction every time step.
Thus, we introduce the consistency loss $loss^C$ to improve the predictions' conformity:
\begin{equation}
loss^C_{i, j} =
\begin{cases}
0& sign((\hat{y}^j_i-0.5)*(\hat{y}^{j-1}_i-0.5))>=0, \\
1& else,
\end{cases}
\end{equation}
where $0.5$ is the decision boundary of positive (malicious) and negative (regular). 

To accelerate the prediction speed, we need to decrease the survival probability as soon as possible.
Thus, we introduced an earliness loss $loss^E$. 
Every time step, the survival probability for positive samples should be as small as possible.
The negative samples' survival probabilities should be as large as possible.
For address $i$ at time split $t$, $loss^E$ is defined as:
\begin{equation}
loss^E_{i,t} =
\begin{cases}
S_{i}(t)& l^i=1, \\
-S_{i}(t)& l^i=0,
\end{cases}
\end{equation}
where $l^i$ is the label of address $i$.

However, the model is hard to predict the correct labels at the early stage due to data insufficiency. 
The model can be perturbed by the wrong predictions in the early period. 
Thus, all the loss items are weighted by $\sqrt{t}$.
the overall loss function is defined as:
\begin{equation}
\begin{aligned}
\mathscr{L} = \sum_{t=1}^{t_M}\sum_{i=1}^{N}\sqrt{t}(loss^P_{i,t} +\gamma_{1}loss^V_{i,t} +\gamma_{2}{loss^C_{i,t}} +\gamma_{3}{loss^E_{i,t}}),
\end{aligned}
\end{equation}
where $\gamma_{1}$ to $\gamma_{3}$ are coefficients to control the contribution between $loss^P$, $l_b$, $loss^C$ and $loss^E$. 
$t_M$ is the time span of training data, $N$ is the training address number.

\section{Experiment and Analysis}
\label{sec:experiment}

\subsection{Data Preparation}
\label{sec:data}

\noindent\textbf{Raw Data and Label Collection}
For higher high credibility, we only select data verified by many participants. 
Thus, we obtained all the data from the $1$-st block to the $610637$-th block (the first block of 2020).
To get the labels for three different types of malicious addresses, namely, hack (hack exchanges and steal tokens), ransomware (encrypt victims' data and demand ransoms in BTC), and darknet (commercial website's address operates via darknets such as I2P).
We performed a manual search on public forums, datasets, 
and prior studies \cite{E2_}, \cite{E3_}, and \cite{R15_}.
For regular addresses, we collected four types of addresses as ``negative samples'', namely exchange, mining, merchant, and gambling.
The negative dataset is also augmented as prior studies~\cite{R15_, E4_, PU_Learn_1, PU_Learn_2}.
The numbers of positive, negative, positive/negative ratios, and dynamic segmentation numbers (within one day) for each malicious type are shown in Table.~\ref{tab:addr_num}.

\begin{table}
    \caption{Dataset Statistics.}
    \begin{center}
    \fontsize{10}{11}\selectfont
    \renewcommand{\arraystretch}{1.2}
    \begin{tabular}{ccccc}
    \toprule
    Type  & Positive &Negative & Posi./Nega. Ratio &Segment \cr
    \midrule
    Hack         & 341      & 79,765   & 0.46\%   &10\cr
    Ransomware   & 1,903    & 50,617   & 3.76\%   &17\cr
    Darknet      & 7,696    & 89318    & 8.62\%   &18\cr
    \bottomrule
    \end{tabular}
    \label{tab:addr_num}
    \end{center}
    \vspace{-2ex}
\end{table}

\subsection{Settings and Metrics}
\label{sec:metrics}
Generally, according to public reports, relevant agencies can detect most malicious behaviors within a day, 
such as hacking or malicious attacks. 
Therefore, in our experiment, to train a model that gives an early warning, we use address first $24$ hours with a $1$ hour interval as training data.
The max time spans for the long-term (LT) and short-term (ST) paths are one week and one day, respectively. 
Notice that the real timespan for a forward path is the smaller one between observing timestep and the pre-defined max timespan, as we can not foresee future data.
We set $0.5$ and $0.01$ as the thresholds for LT and ST paths, respectively, as LT paths aim to find the most critical transaction pairs, and ST paths aim to encode more transaction structure information. 

We average the metrics along the timeline to evaluate the performance. 
The selected metrics are accuracy (Acc.), precision (Prec.), and recall (Rec.).
Besides, the model should predict correct labels fast to prevent economic loss earlier.
Also, due to data insufficiency, the model may predict conflict labels at different timesteps, thus confusing users. 
Thus we require the predictions to be consistent.
To evaluate the earliness and the consistency of the prediction,
we introduce the early-weighted F1 score $F1^{E}$ and consistency-weighted score $F1^{C}$ as follows:
\begin{equation}
\begin{aligned}
\mathit{F1^{E}}& = \frac{\sum_{i=1}^{N}{F1_{i}/\sqrt{i}}}{\sum_{i=1}^{N}{1/\sqrt{i}}}, \\
\mathit{F1^{C}}& = \frac{\sum_{i=1}^{N-1}{{\sqrt{i}}\times{F1_{i}}\times{\mathds{1}_{y_c}(y_{i})}}}{\sum_{i=1}^{N-1}{{\sqrt{i}}}},
\end{aligned}
\end{equation}
where $i$ is the time split index, $y_c$ is the prediction set where current prediction $y_i$ is consistent with the next prediction $y_{i+1}$.
The indicator function $\mathds{1}_{y_c}(y_{i})=1$ when $y_i \in y_c$. $F1_i$ is the $F1$ score of the prediction at the $i$-th time split.

\subsection{Comparison with State-of-the-art Models}
\label{sec:feat_compare}
\begin{table}[htbp]
\caption{Scores of the different prediction models. Inten-M(+Idx) are our \emph{Intention Monitor} with the intention index embedding. 
Underline stands for the best score in the group, 
bold stands for the best score in this dataset.}
\vspace{-2ex}
\begin{center}
\fontsize{8}{9}\selectfont  
\renewcommand{\arraystretch}{1.}
\begin{tabular}{cccccccc}
\toprule
Dataset &Group & Model & $Accuracy$ & $Precision$ & $Recall$ & $F1^E$ & $F1^C$ \cr
\midrule
\multirow{15}{*}{{Hack}} 
                    &\multirow{3}{*}{{Decision Tree}} 
                    &DT            &\ud{0.996}      &0.247        &0.051        &0.084        &0.084\cr
                    & &RF          &\ud{0.996}      &\ud{0.718}   &\ud{0.134}   &\ud{0.238}   &\ud{0.205}\cr
                    & &XGB         &0.992           &0.081        &0.044        &0.049        &0.048\cr
                    \cmidrule(l{0em}r{0em}){2-8} 
                    &\multirow{3}{*}{{Address Graph}}
                    &GCN           &0.736           &0.106        &0.282        &0.163        &0.104\cr
                    & &Skip-GCN    &0.651           &0.143        &\udb{0.524}  &\ud{0.226}   &0.125\cr
                    & &Evo-GCN     &\ud{0.760}      &\ud{0.145}   &0.335        &0.196        &\ud{0.146}\cr
                    \cmidrule(l{0em}r{0em}){2-8}
                    &\multirow{5}{*}{{Sequential Model}}
                    &GRU           &0.970           &0.090        &\ud{0.499}   &0.133        &0.152\cr
                    & &M-LSTM      &0.977           &0.113        &0.445        &0.152        &\ud{0.183}\cr
                    & &CED	       &\ud{0.980}      &\ud{0.115}   &0.401        &\ud{0.154}   &0.181\cr
                    & &SAFE        &0.974           &0.076        &0.394        &0.133        &0.119\cr
                    & &Transformer &0.971           &0.094        &0.464        &0.132        &0.157\cr
                    \cmidrule(l{0em}r{0em}){2-8}
                    &\multirow{4}{*}{{Intention Monitor}}
                    &Status-XGB    &0.996         &\udb{1.000}   &0.257      &0.372        &0.390\cr
                    & &Action-XGB &0.996         &\udb{1.000}   &0.233      &0.370        &0.387\cr
                    & &Inten-M    &0.996         &\udb{1.000}   &0.274      &0.412        &0.440\cr
                    & &Inten-M(+Idx)  &\udb{0.997}   &\udb{1.000}   &\ud{0.298} &\udb{0.436}  &\udb{0.470}\cr

\midrule
\multirow{15}{*}{{Ransomware}} 
                    &\multirow{3}{*}{{Decision Tree}} 
                    &DT            &0.964           &0.073           &0.014        &0.025        &0.019\cr
                    & &RF          &0.964           &0.004           &0.000        &0.000        &0.000\cr
                    & &XGB         &\ud{0.968}      &\ud{0.455}      &\ud{0.421}   &\ud{0.437}   &\ud{0.415}\cr
                    \cmidrule(l{0em}r{0em}){2-8}
                    &\multirow{3}{*}{{Address Graph}}
                    &GCN           &0.878           &0.223           &\ud{0.923}   &0.360        &0.359\cr
                    & &Skip-GCN    &\ud{0.881}      &\ud{0.226}      &0.910        &\ud{0.364}   &\ud{0.361}\cr
                    & &Evo-GCN     &0.866           &0.200           &0.871        &0.322 	     &0.326\cr
                    \cmidrule(l{0em}r{0em}){2-8}
                    &\multirow{5}{*}{{Sequential Model}}
                    &GRU           &0.901           &0.280           &0.856        &0.389        &0.355\cr
                    & &M-LSTM      &0.919           &0.332           &\udb{0.870}  &0.443        &0.418\cr
                    & &CED	       &0.921           &0.329           &0.846        &0.442        &0.415\cr
                    & &SAFE        &0.885           &0.246           &0.856        &0.382        &0.294\cr
                    & &Transformer &\ud{0.928}      &\ud{0.358}      &0.853        &\ud{0.467}   &\ud{0.446}\cr
                     \cmidrule(l{0em}r{0em}){2-8}
                    &\multirow{4}{*}{{Intention Monitor}}
                    &Status-XGB    &0.987         &0.906         &0.724      &0.790        &0.790\cr
                    & &Action-XGB &0.987         &0.910         &0.719      &0.791        &0.780\cr
                    & &Inten-M    &0.986         &\udb{0.930}   &0.770      &0.797        &0.801\cr
                    & &Inten-M(+Idx)  &\udb{0.988}  &0.889           &\ud{0.793}   &\udb{0.820}   &\udb{0.824}\cr

\midrule
\multirow{15}{*}{{Darknet}} 
                    &\multirow{3}{*}{{Decision Tree}} 
                    &DT              &0.908        &0.183        &0.012        &0.020        &0.018\cr
                    & &RF            &0.909        &0.468        &0.013        &0.015        &0.016\cr
                    & &XGB           &\ud{0.913}   &\ud{0.571}   &\ud{0.203}   &\ud{0.291}   &\ud{0.278}\cr
                    \cmidrule(l{0em}r{0em}){2-8} 
                    &\multirow{3}{*}{{Address Graph}}
                    &GCN             &\ud{0.841}   &\ud{0.322}   &0.813        &\ud{0.463}   &\ud{0.465}\crcr
                    & &Skip-GCN      &0.780        &0.260        &\udb{0.889}  &0.400        &0.404\cr
                    & &Evo-GCN       &0.737        &0.216        &0.822        &0.342        &0.343\cr
                    \cmidrule(l{0em}r{0em}){2-8}
                    &\multirow{5}{*}{{Sequential Model}}
                    &GRU             &0.874        &0.518        &0.819        &0.601        &0.438\cr
                    & &M-LSTM        &\ud{0.880}   &\ud{0.528}   &0.822        &0.612        &\ud{0.449}\cr
                    & &CED	         &0.861        &0.490        &0.835        &0.582        &0.414\cr
                    & &SAFE          &0.828        &0.418        &\ud{0.865}   &0.557        &0.343\cr
                    & &Transformer   &0.876        &0.515        &0.856        &\ud{0.615}   &0.438\cr
                     \cmidrule(l{0em}r{0em}){2-8}
                    &\multirow{4}{*}{{Intention Monitor}}
                    &Status-XGB        &0.940         &0.796   &0.694      &0.720        &0.608\cr
                    & &Action-XGB     &0.941         &0.797   &0.710      &0.727        &0.615\cr
                    & &Inten-M        &\udb{0.945} &\udb{0.851}  &0.689        &0.736         &0.624\cr
                    & &Inten-M(+Idx)  &\udb{0.945} &0.794        &\ud{0.766}   &\udb{0.762}   &\udb{0.631}\cr

\bottomrule
\end{tabular}
\vspace{-2ex}
\label{tab:model_compare}
\end{center}
\end{table}

To demonstrate the validity of the temporal information, we compare decision tree models, 
namely Decision Tree (\textbf{DT})~\cite{R12_3}, 
Random Forest (\textbf{RF})~\cite{R12_1}, 
and \textbf{XGB}~\cite{R12_4}.
Then we compare the address graph-based models to justify the ineffectiveness of most existing graph-based for early detection.
Namely \textbf{GCN}, \textbf{Skip-GCN}, and \textbf{Evolve-GCN} in the reference~\cite{TKA9_}.
To verify the validity of our prediction model, we compare four sequential-based models applied in the ``Early Rumor Detection'' task, namely \textbf{GRU}~\cite{E8_}, \textbf{M-LSTM}~\cite{E6_}, \textbf{SAFE}~\cite{E5_}, \textbf{CED}~\cite{E7_}, and \textbf{Transformer}~\cite{R12_5}. 
For our \textbf{Intention Monitor}, $+Idx$ means we replace $z_{i, j}$ with $[z_{i, j} || \mbox{Emb}^I(I^{idx}_{i, j})]$ in Alg~\ref{algo:lstm} and Alg~\ref{algo:attention}.
${Emb}^I$ is the learnable embedding layer for intention index $I^{idx}_{i, j}$ as mentioned in Section~\ref{sec:Intention_fusion}.

First, as shown in Table \ref{tab:model_compare}, our $Inten-M$ model achieves the best performances across all three datasets.
The great improvements come from the effective features and our Intention-VAE module. All these compared models also achieve far better performance with our path features and DT-SC module, which will be discussed later.

For all traditional decision tree algorithms, they do not perform well on the three datasets. 
Because these algorithms are difficult to encode temporal information, 
it is difficult for decision-tree-based machine learning algorithms to consider shifts in the feature decision boundary.
Also, the address features can not provide efficient information for more accurate prediction,
thus, even using decision trees as the backbone, our model still improves $F1^E$ and $F1^C$ significantly compared to the best decision-tree-based model.
For address graph methods, as \cite{E9_} implies, the Address-GCN may lead to over-smoothing issues and the dilution of the minority class. 
In our cases, most neighbors of malicious nodes are victims or shadow addresses. 
In addition, since the transaction network is usually small in the early stage of the address, there are many shadow addresses in this network, which makes the Address-GCN models challenging to obtain valuable transaction pattern information. Thus the models do not perform well. 

Rather than focusing on the address transaction graph's structure, sequential models encode the temporal pattern more directly. 
As shown in Table.~\ref{tab:model_compare}, compared to the best Address-GCN model, 
they have an average improvement of $9.76\%$ and $15.78\%$ in $F1^E$ and $F1^C$ on the three datasets.
This improvement verifies the effectiveness of temporal patterns.
However, malicious behavior generally involves many transactions, which are often not directly related to the target address. For example, the malicious address will use many shadow addresses for transit. Therefore, the address feature cannot reflect the actual attributes of the address in time, while our backward path feature can describe how the funds flow into the address.
Similarly, the malicious address may also transfer funds in a certain way. The most common method is the peeling chain.
Our forward path feature can provide relevant information, which is a challenging task for address features.
As shown in Table.~\ref{tab:model_compare}, 
with the intention attention module, the model can finetune the predictions of $Status$-$XGB$ and $Action$-$XGB$,
thus achieving better performance across all three datasets.
Compared to the best sequential model,
our $Inten$-$M$ model can achieve an average improvement of $93.68\%$ and $94.04\%$ on $F1^C$ and $F1^E$,
proving the great effectiveness of our path feature and corresponding feature selection.
Also, the encoding of the binarized Intention index improves the model's performance one step further, 
which justifies the boundary in the Intention-VAE module is useful for prediction.

\subsection{Feature Combination and Selection}
\label{sec:predict_ana}
In this subsection, we evaluate the effectiveness of our path features and feature selection scheme.
We also justify the generality with compatible models.

\noindent\textbf{Feature Combination}.
To verify the effectiveness of path features, we compare three decision tree models, 
namely the address feature (\textbf{AF}) model, the long-term path features (\textbf{+LT}) model, and the short-term path features (\textbf{+ST}) model.
\textbf{+X} means add feature \textbf{X} to previous model.
As shown in Table \ref{tab:feat_compare}, \textbf{AF} performed poorly on the three datasets. 
Path features significantly improve the performance for all metrics.
We speculate that most malicious addresses require victims to transfer money once they are created. 
It makes them similar to exchange or financial service addresses in the early stage.
For example, exchange or merchant services will also create new addresses for security. 
As a result, \textbf{AF} model can only find those extremely abnormal addresses, which results in poor performance. 

\noindent\textbf{Feature Selection}.
Feature selection is crucial for the model's generalization ability.
To justify the effectiveness of our feature selection scheme, 
we also compare the \textbf{T/C} model with the one without a selection scheme \textbf{+ST}. 
Trimming scheme (\textbf{T}) stands for trimming off features in the deletion list, 
Complement scheme (\textbf{C}) stands for complement features in the complement list. 
As shown in Table~\ref{tab:feat_compare}, feature selection significantly improves the performances of decision trees across all the datasets. 
The $F1^{C}$ and $F1^{E}$ are enhanced by an average of $13.88\%$ and $9.33\%$, respectively. 
The major improvement comes from the \emph{recall} score. 
Since malicious activities behave abnormally in various aspects, their most significant features are also different. 
Through our automatic feature selection method, models can fully use the powerful path features. 
Also, they can adapt to different malicious activities easily, mitigate input noise, 
and significantly reduce the workload of manual selection.

Since our feature selection scheme is based on the decision tree model, 
we implement them on other compatible models to verify the generalization of path features and the feature selection scheme.
As shown in Table~\ref{tab:feat_compare}, our path feature and feature selection scheme improve the performance for all models. Especially for the decision tree model, the augmented \textbf{XGB} model even outperforms most sequential models.
\begin{table}[htbp]
\caption{Scores of different features (address feature (\textbf{AF}), long-term path features (\textbf{+LT}), short-term path features (\textbf{+ST}), and selection schemes(trimming/complement scheme (\textbf{T/C}). $\Delta$ stands for performance differentiation after applying path features and selection schemes.)}
\begin{center}
\fontsize{8}{9}\selectfont  
\renewcommand{\arraystretch}{1.}
\begin{tabular}{ccccccc}
\toprule
Dataset & Model & $Accuracy$ & $Precision$ & $Recall$ & $F1^E$ & $F1^C$ \cr
\midrule
\multirow{11}{*}{{Hack}} 
                &DT(AF)   &0.996   &0.247   &0.051   &0.084   &0.084\cr
                &DT(+LT)  &0.995   &0.221   &0.053   &0.079   &0.092\cr
                &DT(+ST)  &0.996   &0.882   &0.116   &0.249   &0.278\cr
                &DT(+T/C) &0.996   &0.821   &0.155   &0.259   &0.277\cr
                \cmidrule(l{0em}r{0em}){2-7} 
                &RF($\Delta$)          &+0.000   &+0.282   &+0.053   &+0.095   &+0.093\cr
                &XGB($\Delta$)         &+0.004   &+0.919   &+0.181   &+0.291   &+0.333\cr
                &GRU($\Delta$)         &+0.018   &+0.103   &-0.134   &+0.089   &+0.105\cr
                &M-LSTM($\Delta$)      &+0.014   &+0.184   &-0.116   &+0.114   &+0.134\cr
                &CED($\Delta$)         &+0.006   &+0.044   &-0.050   &+0.037   &+0.040\cr
                &SAFE($\Delta$)        &+0.006   &+0.025   &+0.004   &+0.034   &+0.034\cr
                &Transf.($\Delta$)     &+0.018   &+0.089   &-0.176   &+0.067   &+0.070\cr

\midrule
\multirow{11}{*}{{Ransomware}} 
                &DT(AF)   &0.964   &0.073   &0.014   &0.025   &0.019\cr
                &DT(+LT)  &0.965   &0.313   &0.090   &0.125   &0.126\cr
                &DT(+ST)  &0.967   &0.592   &0.270   &0.354   &0.330\cr
                &DT(+T/C) &0.968   &0.616   &0.399   &0.463   &0.417\cr
                \cmidrule(l{0em}r{0em}){2-7} 
                &RF($\Delta$)          &+0.005   &+0.926   &+0.269  &+0.394   &+0.425\cr
                &XGB($\Delta$)         &+0.014   &+0.334   &+0.283  &+0.295   &+0.313\cr
                &GRU($\Delta$)         &+0.047   &+0.169   &-0.027  &+0.146   &+0.186\cr
                &M-LSTM($\Delta$)      &+0.027   &+0.109   &-0.038  &+0.087   &+0.113\cr
                &CED($\Delta$)         &+0.022   &+0.092   &-0.026  &+0.070   &+0.095\cr
                &SAFE($\Delta$)        &+0.030   &+0.065   &-0.020  &+0.076   &+0.073\cr
                &Transf.($\Delta$)     &+0.023   &+0.102   &-0.091  &+0.067   &+0.089\cr
                 
\midrule
\multirow{11}{*}{{Darknet}} 
                &DT(AF)   &0.908   &0.183   &0.012   &0.020   &0.018\cr
                &DT(+LT)  &0.911   &0.593   &0.099   &0.143   &0.140\cr
                &DT(+ST)  &0.921   &0.609   &0.404   &0.484   &0.452\cr
                &DT(+T/C) &0.922   &0.618   &0.446   &0.517   &0.461\cr
                \cmidrule(l{0em}r{0em}){2-7} 
                &RF($\Delta$)          &+0.006   &+0.324   &+0.428   &+0.503   &+0.499\cr
                &XGB($\Delta$)         &+0.036   &+0.206   &+0.417   &+0.380   &+0.349\cr
                &GRU($\Delta$)         &+0.041   &+0.112   &+0.032   &+0.091   &+0.094\cr
                &M-LSTM($\Delta$)      &+0.036   &+0.106   &+0.036   &+0.084   &+0.085\cr
                &CED($\Delta$)         &+0.053   &+0.138   &+0.002   &+0.104   &+0.118\cr
                &SAFE($\Delta$)        &+0.046   &+0.087   &-0.030   &+0.060   &+0.081\cr
                &Transf.($\Delta$)     &+0.035   &+0.101   &+0.001   &+0.071   &+0.084\cr  

\bottomrule
\end{tabular}
\vspace{-2ex}
\label{tab:feat_compare}
\end{center}
\end{table}

\subsection{Effect of Status\&Action Cluster Number}
\label{sec:cluster_num_ana}

\begin{figure}
	\centering
	\vspace{-0ex}
	\includegraphics[width=1.\columnwidth, angle=0]{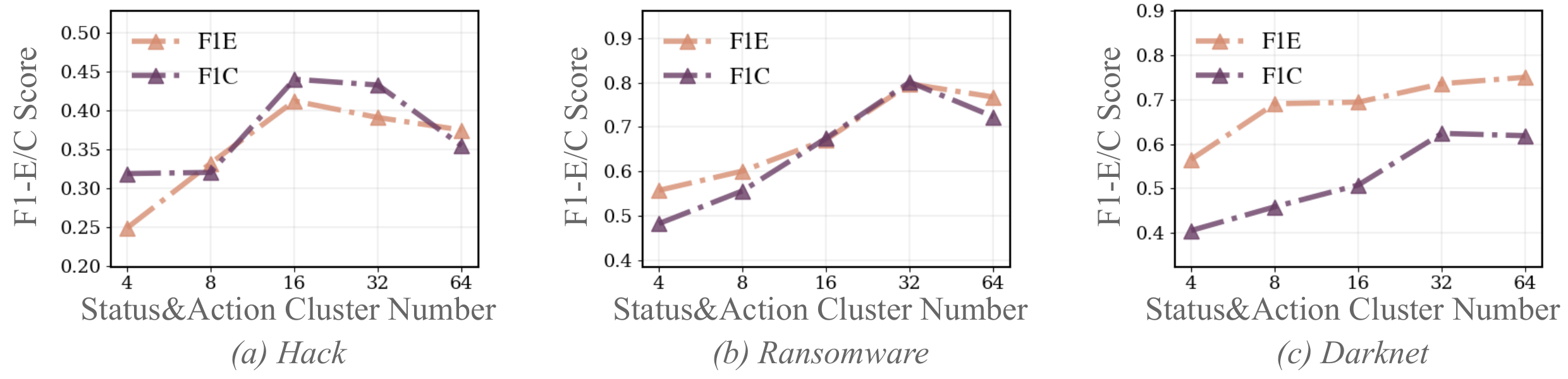}
	\vspace{-4ex}
	\caption{\emph{$F1^E$} and \emph{$F1^C$} of different status and action cluster number.}
	\label{fig:cluster_num_evolve}
	\vspace{-2ex}
\end{figure}
The cluster numbers of status and action represent the granularity of address status and behavior.
If the cluster number is too small, it will limit the model's ability to distinguish and extract adequate information.
On the contrary, if the cluster number is too large, it will introduce outliers and deteriorate the model's interpretability. 
To this end, we analyze the effect of cluster number on model performance. We test $5$ cluster numbers ($4$,$8$,$16$,$32$,$64$), and the results are shown in Figure ~\ref{fig:cluster_num_evolve}.

As expected, models with larger cluster numbers can distinguish more states at the beginning and perform better.
However, when the cluster number increases, the model may introduce more redundant noise, leading to poor performance.
This phenomenon is particularly evident in the hack dataset.
This is because compared to other malicious types, hack addresses tend to have fewer actions, so the diversity of their states and actions will be less. Therefore, the best cluster number is $16$, and the performance gradually deteriorates afterward.
Moreover, because ransomware and darknet have more operations or requirements for users, the optimal cluster number will be larger. For ransomware, the best cluster number is 32 because it performs best.
For darknet, we also choose $32$ as the best cluster number because the performance difference between $32$ and $64$ is marginal. And the model with a smaller cluster number has better interpretability.

\subsection{Status Actions and Intention}
\label{sec:stat_and_inten}
\begin{figure}
	\centering
	\vspace{-0ex}
	\includegraphics[width=1.\columnwidth, angle=0]{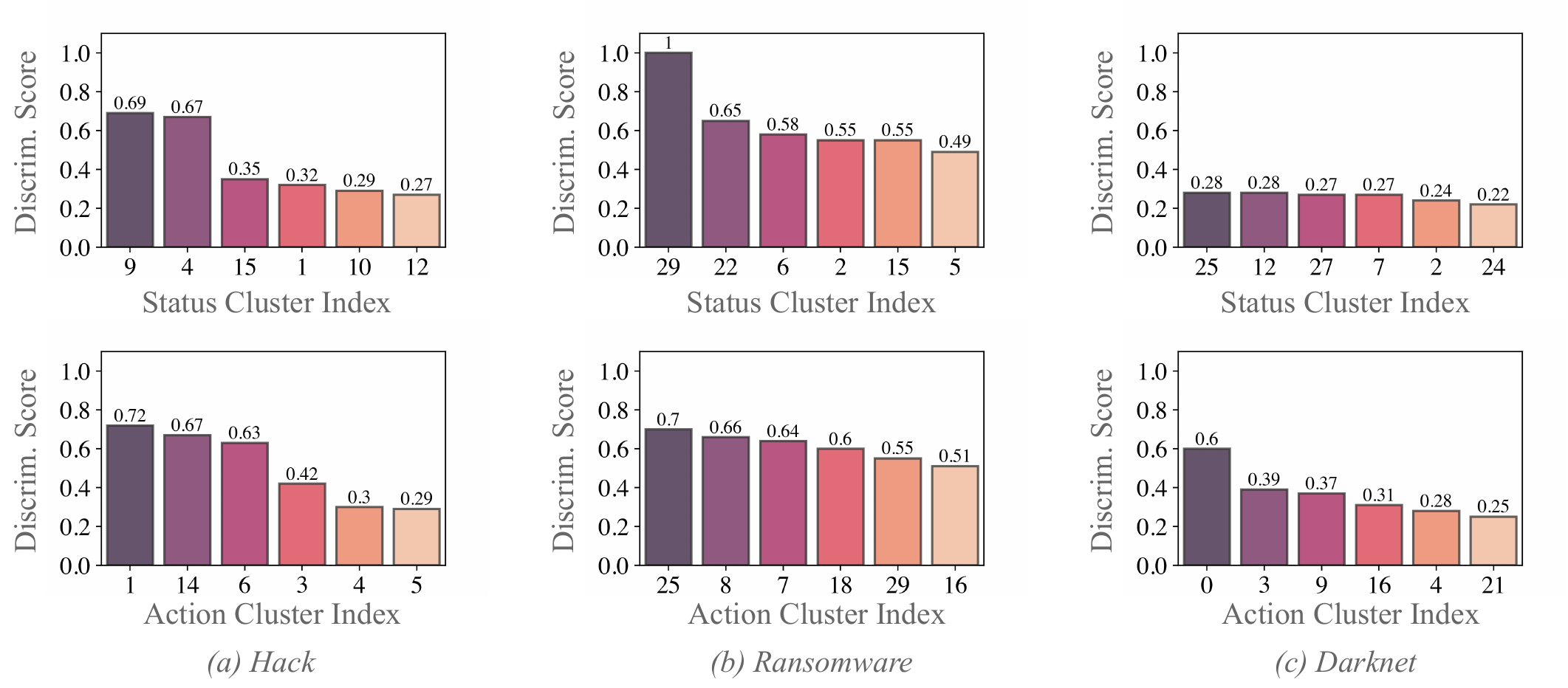}
	\vspace{-4ex}
	\caption{Top-6 differentiated status and action. Labels on the x-axis stand for the index of status and 
                 action. The value above each bar is the corresponding discrimination score.}
	\label{fig:stat_prop_diff}
	\vspace{-2ex}
\end{figure}

\noindent\textbf{Differentiation Analysis}.
We select the top six with the largest differences between malicious and regular addresses.
As shown in Fig.~\ref{fig:stat_prop_diff}, the x-axis represents the cluster index of status and action. 
The y-axis is the discrimination score (the proportion difference between the positive and negative samples divided by the sum of the proportion). A higher score means the cluster appears more in a specific class.

We can see that some status and action clusters can be used as effective discriminant indicators for hack and ransomware addresses. 
Especially for status 29 in the ransomware dataset, the discrimination score is 1. That is to say, as long as status 29 appears in the address status sequence, the model can confidently determine the address's label.

As for the darknet address, since its operation method is similar to the typical trading platform,
therefore, the cluster discrimination ability is not as strong as the other two data sets.
However, when we analyze 2-gram and 3-gram (a contiguous sequence of 2 and 3 cluster indices), 
we found that the order in which each cluster appears in status and action can give a higher discrimination score for darknet addresses.
On the one hand, this justifies the importance of temporal information. On the other hand, it shows that our status and action can be well combined with timing analysis to achieve better results.

\begin{figure}
	\centering
	\vspace{-0ex}
	\includegraphics[width=1.\columnwidth, angle=0]{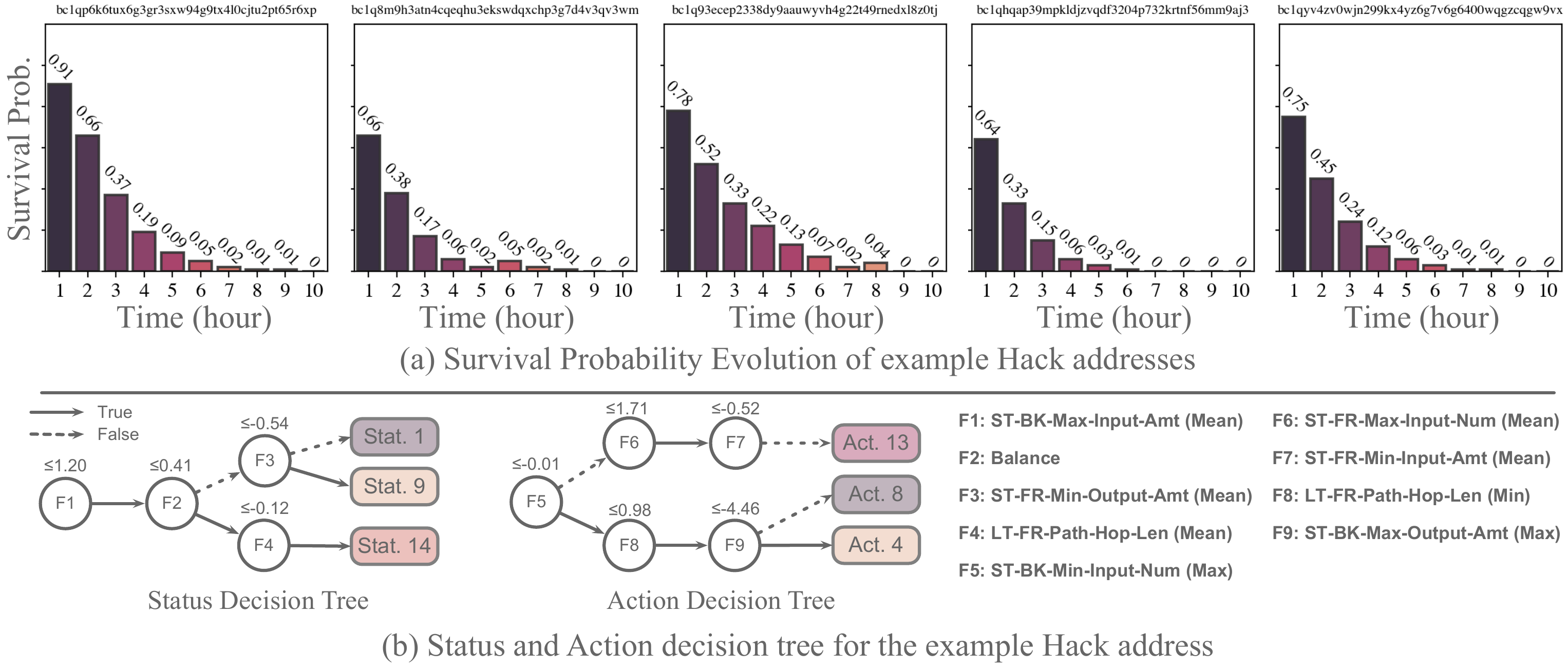}
	\vspace{-4ex}
	\caption{(a) Survival probability evolution of example hack addresses. 
	         (b) Status and action decision tree for the example hack address.}
	\label{fig:hack_status}
	\vspace{-2ex}
\end{figure}

\noindent\textbf{Case Study}.
We use the infamous hack incident of the Binance crypto exchange on May 7, 2019, \footnote{https://www.cnbc.com/2019/05/08/binance-bitcoin-hack-over-40-million-of-cryptocurrency-stolen.html} in which hackers stole over 7000 BTCs worth of 40 million, as a case study to illustrate our model's capability to interpret prediction result and offer valuable insights into the malicious behavior. 

As shown in Fig.~\ref{fig:hack_status} (a), 
our method successfully detects the 5 sample hacker addresses involved by the end of the first $9$ hours since its creation, which is \emph{12 hours before the stolen BTCs were transferred away}. 
We quote from a Binance statement that "It was unfortunate that we were not able to block this withdrawal before it was executed.", an inevitable tragedy with only retrospective analysis but totally preventable with our early detection.   

To interpret our prediction result, we first zoom into one of the hacker's addresses bc1***6xp\footnote{bc1qp6k6tux6g3gr3sxw94g9tx4l0cjtu2pt65r6xp}. 
The prediction result can be easily interpreted by examining the semantics behind the status and action that form the intention motif. 
By the $16$-th hour, the corresponding status and action sequences are $[1-9-9-9-9-9-9-9-9-9-9-9-9-9-9-14]$ and $[8-8-8-8-8-8-8-8-8-8-8-8-4-8-8-13]$. 
To interpret the semantic meaning of status and action, we build corresponding status and action decision trees.
Take status as an example. After obtaining all the status vectors via clustering, we categorize all the statuses using the decision tree (the number of statuses is equal to the number of categories).
Each status can be interpreted by trailing the corresponding decision tree from the root to the leaf, as shown in Fig.~\ref{fig:hack_status} (b).

Regarding feature 2, the status decision tree judges "when the input volume of each ST-BK path is not very large, whether the balance can reach a certain amount." That is to say, is it in a state of "remitting funds through multiple channels and having enough balance"?
\textbf{Status 1} essentially indicates that the initial balance is very high, and the asset comes from lots of ST-BK paths.
\textbf{Status 9} mainly determines whether the address has transferred out the asset. As shown in Figure\ref{fig:stat_prop_diff}, there is already a relatively high discrimination score for statuses 1 and 9. Our model can give the correct prediction before the end of status 9, which justifies the effectiveness of our status proposer.
Furthermore, at the $16$-th hour, the address status is 14. 
\textbf{Status 14} implies that the funds come through multiple ST-BK paths, but the balance is insufficient. 
The low balance may be because the volumes of ST-BK paths are insufficient or the address transferred out its tokens with short LT-FR paths. 

As for action clusters, 
\textbf{Action 8} is a waiting action, which is used to describe that the features of the address are relatively stable without much change. We can also see from the figure that all features on the decision path will not have too large or too small values.
\textbf{Action 4} is similar to action-8. The main difference is that the address introduces a small amount of ST-BK path, which leads to a significant decrease in the overall corresponding feature value.
\textbf{Action 13} mainly describes the property change in the ST-FR path. 
As shown in Fig.~\ref{fig:hack_status} (b), this action describes that the address introduces the ST-FR path, most of which are single chains, and the transfer amount is relatively large.

\begin{figure}
	\centering
	\vspace{-0ex}
	\includegraphics[width=1.\columnwidth, angle=0]{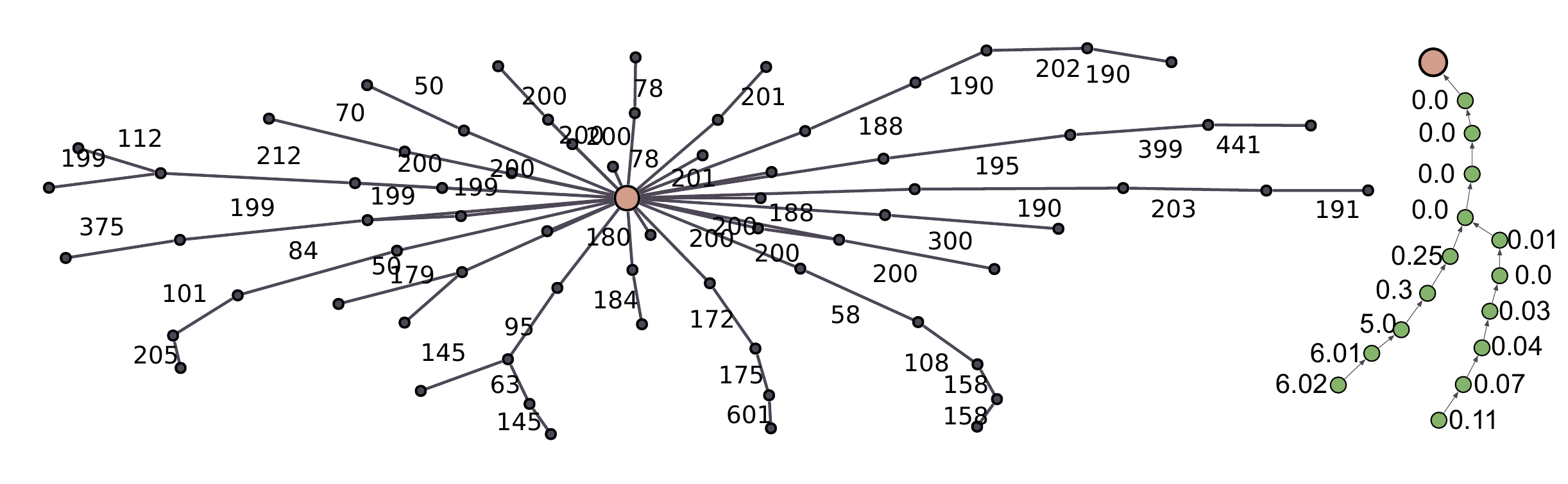}
	\vspace{-2ex}
	\caption{Short-term backward (ST-BK) paths for the instance address in the first $24$ hours (gray and green edges stand for ST-BK paths of the address's first and second incoming transactions.). The numbers stand for the transaction volumes. The pink nodes are instance-related transactions.}
	\label{fig:case_ST_BK}
	\vspace{-2ex}
\end{figure}

In reality, this malicious address received a transfer of $555.997$ BTCs at creation through $71$ input transitions, with no output transactions. 
The status sequence can also be observed in the related transaction\footnote{e8b406091959700dbffcff30a60b190133721e5c39e89bb5fe23c5a554ab05ea}, 
in which the hacker manipulated Binance's address and divided it into $71$ inputs, each containing $100$ BTCs. 
This justifies the "Multi ST-BK Paths" property of Statuses 1 and 9.
The pattern can be seen in Fig.~\ref{fig:case_ST_BK}. 
Those black edges are the ST-BK paths related to the first input transfer.
Besides, the ``waiting period'' is represented by action 8.

By the end of $13$-th hour, it received another transfer of about $0.00008631$ BTC.
This tiny transaction will not change the address's status,
but it can be reflected by action 4 as this tiny transaction introduces an ST-BK path.
After the transaction of a tiny amount, there is a bulk transfer of address's all BTC at the $16$-th hour.
By analyzing the LT-FR paths during the $16$-th hour, we found the path hop lengths are also lower than 2.
Also, the introduced LT-FR paths can be reflected by action 13.
The whole evolution justifies consistency between status and the address's real state.

Moreover, valuable insights can be acquired from our intention monitor. 
For example, the extremely tiny amount of $0.00008642$ BTC received by the malicious address by the end of $13$-th hour is highly likely the corroborating evidence that it is a trial transfer to test whether the transfer operation is successful, as a specific signal transaction to coordinate and synchronize multiple addresses' hacking operations automatically. This was also validated by Binance's statement in which they pointed out that “The hackers had the patience to wait and execute well-orchestrated actions through multiple seemingly independent accounts at the most opportune time.”

As another example, although our model has already given the prediction in the $9$-th hour, combined with the subsequent status of our proposed ST-BK path, we can even identify potentially a group of hackers for this hacking incident.
As shown in Fig.~\ref{fig:case_ST_BK}, the green edge is a signal transaction after $13$ hours, and the amount on it is tiny but introduces one LT-BK path and two ST-BK paths. These two ST-BK paths merged into a single track before importing to the hack address.

We back-tracked the source of the ST-BK path of the signal transaction. We found that the $21$ hack addresses that participated in this hacking incident were linked through the signal transactions. 
Even more surprisingly, they have the same source coming from the address\footnote{1GrdXZpyBfiNSVereX5t5UQRHfeh192Cc6}. Our belief that multiple addresses launched this hacking and synchronized among themselves through signal transactions again echoes the collaborative schemes claimed in the Binance statement. 

\subsection{Scalability Analysis}
\label{sec:scala_ana}
Generally speaking, users only want to monitor the new and large-volume addresses that have transactions with them. 
Those addresses are likely to participate in dangerous activities. 
Our system will dynamically monitor and update their transaction features every hour, as shown in Fig. \ref{fig:asset_trans_path}(b).
Then, the model proposes segments according to the default position and prepares the segment and status representations. 
In this way, our intention monitor can predict the labels dynamically.

In addition, to verify the model's scalability, we randomly selected $1,000$ blocks (from the first block of $2020$ to the first block of $2022$) and collected the daily BTC price during this period.
We filter out transactions lower than \$$10,000$ and retrieve addresses with a lifespan smaller than one week. 
We got $299,767$ transactions and $181,221$ addresses.
To get stronger proof of the model's scalability, we fetch the data of the early $200$ hours (larger than one week) with a one-hour interval. 
The time costs under single-process and multi-process (acceleration by 20 processes) are illustrated as follows:
\begin{table}
\centering
\fontsize{8}{9}\selectfont 
\caption{Time cost of different input data, including block number, transaction number, address number), address feature, LT transfer path, and ST transfer path.}
\vspace{-1ex}
\begin{tabular}{ccccccccc}
\toprule
Block Number           &Transaction Number      &Address Number      &Address Feature    &LT-Path   &ST-Path\cr
\midrule 
1,000 (Single-Process)    & 299,767  & 181,221    & 0.51h    & 72.53h    &130.33h\cr
1,000 (Multi-Process)     & 299,767  & 181,221    & 0.03h    & 3.82h     &6.86h\cr
\midrule 
Avg (Multi-Process)       & 300      & 181        & 0.11s    & 13.75s    &24.70s\cr
\bottomrule
\end{tabular}
\vspace{-2ex}
\label{tab:time_cost}
\end{table}

On average, for the early $200$ hours, preparing an address feature takes $0.11$s.
The LT and ST path feature take $13.75$s and $24.70$s, respectively.
So we only need $2.42$s to prepare data in every one-hour interval.



\section{Conclusion}
\label{sec:conclusion}
This paper presents Intention Monitor, a novel framework for the early detection of malicious addresses on BTC. 
After proposing two kinds of asset transfer paths, 
we select, complement, and split the feature sequence for different malicious activities with a decision tree based strategy. 
In particular, we propose status and action vectors to describe the temporal behaviors and global semantic status and action.
We build the Intention-VAE to propose intent-snippets and weight the contribution of status and action backbone predictions dynamically. A survival module based on Intention-VAE fine-tunes the weighted predictions and groups intent-snippets into the sequence of intention motif. 
We quantitatively and qualitatively evaluated the model on three malicious address datasets.
Extensive ablation studies were conducted to determine the mechanisms behind the model's effectiveness. 
The experimental results show that the proposed method outperformed the state-of-the-art baseline approaches on all three datasets.
Furthermore, a detailed case study on Binance Hack justifies that our model can not only explain suspicious transaction patterns but can also find hidden abnormal signals.

\section{Acknowledgments}
This research is supported by the 
National Research Foundation Singapore under its 
Emerging Areas Research Projects (EARP) Funding Initiative, 
Industry Alignment Fund – Pre-positioning (IAF-PP) Funding Initiative, 
the NSFC (72171071, 72271084, 72101079),  
and the Excellent Fund of HFUT (JZ2021HGPA0060). 
Any opinions, findings and conclusions or recommendations expressed in this material are those of the author(s) and do not reflect the views of National Research Foundation, Singapore.

\bibliographystyle{ACM-Reference-Format}
\bibliography{acmart}

\end{document}